\documentclass[10pt,twocolumn,letterpaper]{article}

\usepackage{cvpr} 
\usepackage{multirow}
\usepackage{booktabs}

\usepackage{graphicx}
\usepackage{amsmath}
\usepackage{amssymb}
\usepackage{booktabs}
\usepackage{blindtext}

\def\paperID{4504} 
\def\confName{CVPR}
\def\confYear{2025}

\title{GoLF-NRT: Integrating Global Context and Local Geometry for Few-Shot View Synthesis\thanks{This work is supported by National Key R\&D Program of China under Grant No. 2021YFF0900500, and the National Natural Science Foundation of China under Grant No. 62001432.}}

\author{
You Wang$^{1}$\thanks{You Wang and Li Fang contributed equally to this work. Li Fang is the corresponding author.} \quad
Li Fang$^{1}\footnotemark[2]$ \quad
Hao Zhu$^{2}$ \quad
Fei Hu$^{1}$ \quad
Long Ye$^{1}$ \quad
Zhan Ma$^{2}$ \\
$^{1}$Key Laboratory of Media Audio and Video (Communication University of China), \\
Ministry of Education, Beijing 100024, China \\
$^{2}$School of Electronic Science and Engineering, Nanjing University, Nanjing 210023, China \\
{\tt\small \{wyou621, lifang8902, hufei, yelong\}@cuc.edu.cn} \quad
{\tt\small \{zhuhao\_photo, mazhan\}@nju.edu.cn}
}

\begin{document}
\maketitle
\begin{abstract}
Neural Radiance Fields (NeRF) have transformed novel view synthesis by modeling scene-specific volumetric representations directly from images. While generalizable NeRF models can generate novel views across unknown scenes by learning latent ray representations, their performance heavily depends on a large number of multi-view observations. However, with limited input views, these methods experience significant degradation in rendering quality. To address this limitation, we propose GoLF-NRT: a Global and Local feature Fusion-based Neural Rendering Transformer. GoLF-NRT enhances generalizable neural rendering from few input views by leveraging a 3D transformer with efficient sparse attention to capture global scene context. In parallel, it integrates local geometric features extracted along the epipolar line, enabling high-quality scene reconstruction from as few as 1 to 3 input views. Furthermore, we introduce an adaptive sampling strategy based on attention weights and kernel regression, improving the accuracy of transformer-based neural rendering. Extensive experiments on public datasets show that GoLF-NRT achieves state-of-the-art performance across varying numbers of input views, highlighting the effectiveness and superiority of our approach. Code is available at \href{https://github.com/KLMAV-CUC/GoLF-NRT}{https://github.com/KLMAV-CUC/GoLF-NRT}.
\end{abstract}

\section{Introduction}
\label{sec:intro}

\begin{figure}[t]
  \centering
  \includegraphics[width=0.95\linewidth]{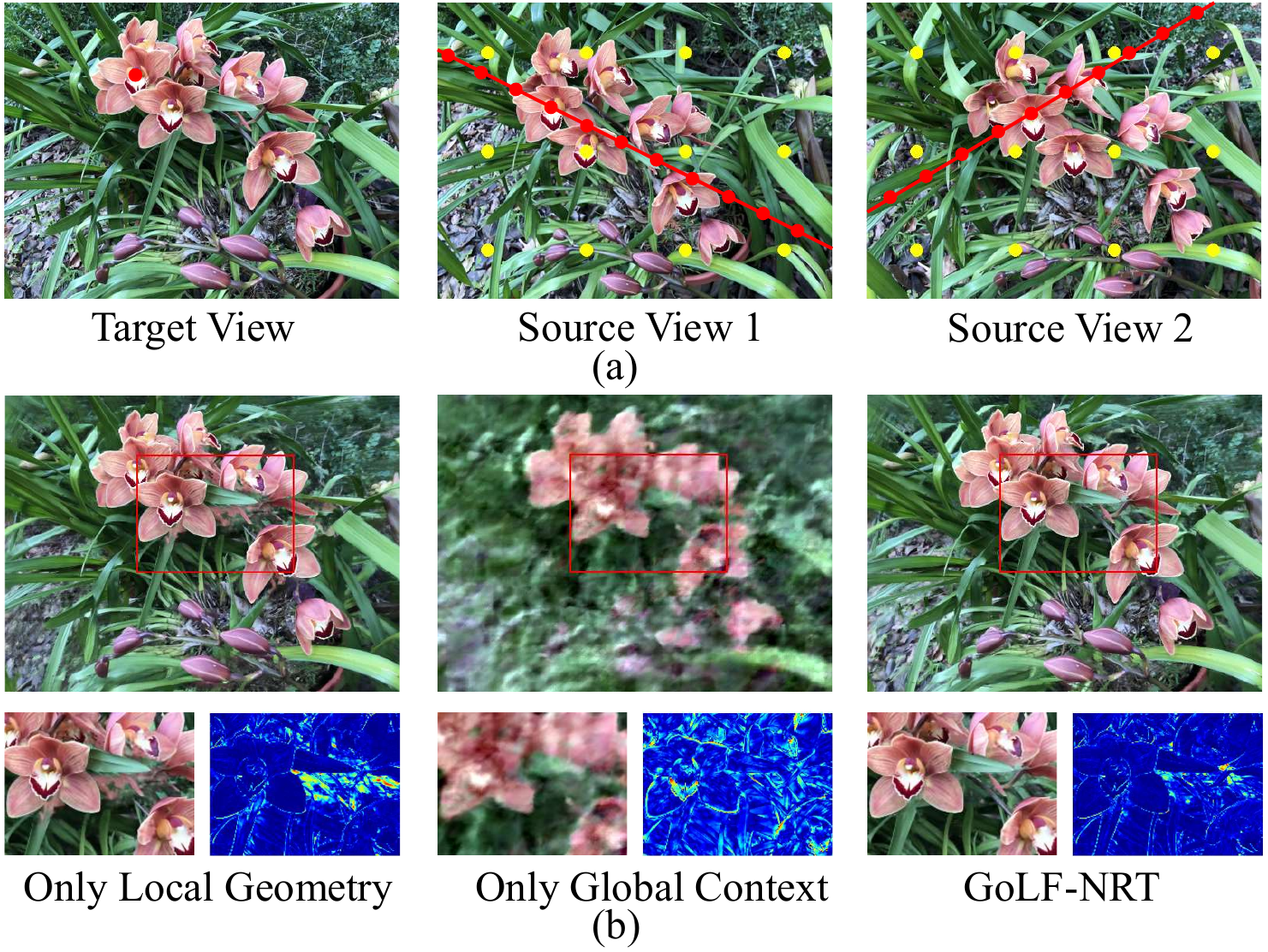}
   \caption{The integration of global context and local geometry in GoLF-NRT significantly enhances reconstruction accuracy. (a) Local geometric features (red) are extracted along the epipolar line of the target pixel, effectively capturing local geometric constraints, while surrounding features (yellow) provide the global context. (b) Relying solely on local features can introduce depth ambiguities and artifacts, while using only global features often fails to capture fine details. By combining both, GoLF-NRT achieves more precise and detailed reconstructions.}
   \label{fig:teaser}
\end{figure}

Neural Radiance Fields (NeRF) \cite{mildenhall2021nerf} and its subsequent refinements \cite{barron2021mip, barron2022mip, barron2023zip, yu2021plenoctrees,sun2024recent} have achieved remarkable success in generating realistic, high-resolution, and view-consistent scenes for novel view synthesis. Despite these advancements, conventional NeRF approaches require expensive optimization with a large number captured images for each scene, hindering their widespread adoption in real-world applications. These constraints underscore the necessity of few-shot generalizable neural rendering, which aims to render unseen scenes from novel viewpoints using a limited number of input views.

To address the generalization issue, existing methods mainly introduce explicit or implicit priors to extract local geometrical features, such as the depth maps \cite{deng2022depth, roessle2022dense}, cost volumes \cite{chen2021mvsnerf, wang2021ibrnet, yu2021pixelnerf}, geometric consistency across views \cite{bao2023and, kwak2023geconerf, niemeyer2022regnerf}, and epipolar features encoded in attention maps \cite{suhail2022light, suhail2022generalizable, varma2022attention}. However, these methods often require many input views to accurately capture geometry. With fewer views, they tend to produce over-smoothed textures or artifacts, particularly in occluded or reflective regions.

To address the few-shot issue, various types of auxiliary information have been incorporated with local geometrical features, including the normalization flow \cite{zhang2021ners}, semantic label constraints \cite{jain2021putting, gao2023surfelnerf} and frequency content \cite{yang2023freenerf}. However, these human-designed auxiliary features can only be reliably extracted with at least 3 views. When the number of input views further decreases, the inherent ambiguities in auxiliary information, such as large motion, incomplete semantic labels and restricted spectral bandwidth, exacerbate the over-smooth textures or artifacts in occlusion boundaries.

In this paper, we present GoLF-NRT (Global Local feature Fusion based Neural Rendering Transformer), a coarse-to-fine model that integrates global context features with enhanced local geometric features. Our model comprises two key modules: the global context feature extraction module and the local geometric feature extraction module. The global context feature extraction module employs a cross-view transformer-based encoder with multi-axis sparse attention to effectively learn the scene representation. This representation is then decoded into a global context feature for each ray, enabling the model to develop a comprehensive understanding of the scene from a limited set of viewpoints. The local geometric feature extraction module further refines this understanding by aggregating features along epipolar lines in each input view, alternating between a view transformer and a ray transformer. Unlike previous methods \cite{zhu2023caesarnerf, venkat2023geometry} that directly fuse geometric and auxiliary features, our model uses the global context feature as a coarse prediction to better evaluate the contributions of different views. This approach ensures consistent scene understanding across views, improving the model's ability to handle depth ambiguity and occlusions. Additionally, we introduce an adaptive sampling strategy that transforms attention weights \cite{vaswani2017attention} into a smooth, regularized probability density function (PDF) using kernel regression \cite{2005Image, 2006Robust, 2007Kernel}, thereby improving the model's geometric perception and overall rendering quality.

Compared to prevailing methods such as \cite{suhail2022generalizable, varma2022attention, min2024entangled, zhu2023caesarnerf}, our GoLF-NRT framework stands out by synthesizing target rays through the integration of global context understanding and local geometric perception. This distinctive combination enables our approach to perform exceptionally well in both many-shot and few-shot scenarios, offering a robust and versatile solution to the challenges of neural rendering across varying input conditions. Our main contributions are summarized as follows:
\begin{itemize}
\item We propose GoLF-NRT, a generalizable neural rendering technique that integrates global context with local geometry, ensuring consistent scene understanding across views and enhances the handling of depth ambiguity and occlusions.

\item We introduce an adaptive sampling strategy using attention mechanisms and kernel regression to extract precise local geometric features, improving the model's ability to accurately perceive geometry.

\item Through extensive experiments and comparisons with state-of-the-art methods, we demonstrate that our framework consistently outperforms current generalizable neural rendering methods across various datasets, particularly in scenarios with limited input data.
\end{itemize}

\begin{figure*}[t]
  \centering
  \includegraphics[width=0.95\linewidth]{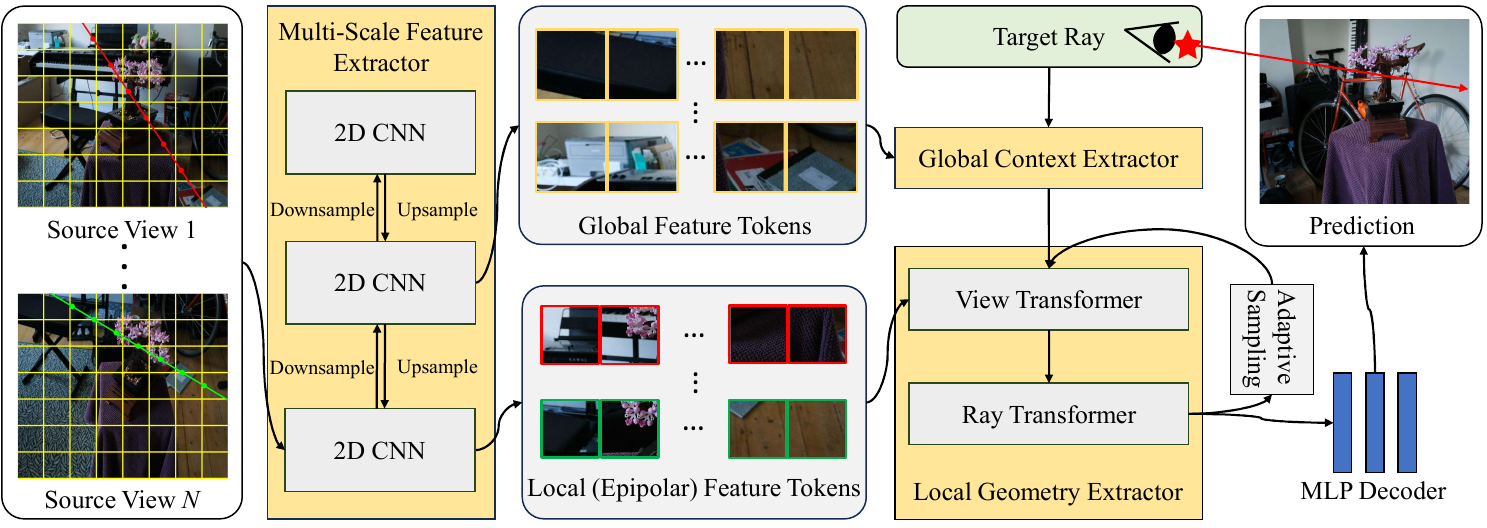}
  \caption{Overview of GoLF-NRT: 1) A FPN extracts multi-scale features from input views. 2) Coarse features are encoded to create scene representations, from which global context features are decoded for each target ray. 3) These global context features query local geometric features along the epipolar lines of all source views. The global and local features are concatenated and processed to predict the final target ray representation. 4) Pixel colors are directly predicted using a MLP.}
  \label{fig:pipeline}
\end{figure*}

\section{Related Works}
\label{sec:rw}
\paragraph{NeRF and Generalizable NeRF.} NeRF \cite{mildenhall2021nerf} reconstructs images by predicting point density and color within a radiance field based on spatial coordinates and viewing directions. Follow-up works have improved rendering quality \cite{barron2021mip, barron2022mip, barron2023zip, hu2023tri}, increased speed \cite{chen2022tensorf, chen2023mobilenerf, hedman2021baking, reiser2023merf, sun2022direct, yu2021plenoctrees}, and enabled editing \cite{liu2021editing, sun2022fenerf, wang2022clip, xiang2021neutex, xu2022deforming}. However, NeRF's scene-specific implicit functions limit generalization. Generalizable NeRFs address this by encoding features at each 3D point and decoding them into volume density and radiance, allowing cross-scene generalization without retraining \cite{yu2021pixelnerf, trevithick2021grf, chen2021mvsnerf, liu2022neural, chen2023explicit, xu2024murf}. Recently, attention mechanisms and transformers have further advanced novel view synthesis and generalization \cite{chibane2021stereo, huang2023local, li2021mine, reizenstein2021common, wang2022generalizable, yang2023contranerf, johari2022geonerf}. IBRNet \cite{wang2021ibrnet} uses transformers for volumetric density prediction, while other methods \cite{cong2023enhancing, kulhanek2022viewformer, suhail2022generalizable, varma2022attention, sajjadi2022scene} employ fully attention-based models for scene representation and ray color estimation. EVE-NeRF \cite{min2024entangled} enhances representation quality by integrating view and epipolar information, though it may struggle in few-shot scenarios due to data limitations and boundary artifacts.

\paragraph{Few-shot Generalizable NeRF.} Various NeRF approaches address the few-shot problem. WaH-NeRF \cite{bao2023and} tackles sample-position confusion with a deformable sampling strategy and mutual information loss. RegNeRF \cite{niemeyer2022regnerf} regularizes color predictions for unseen viewpoints using a pre-trained flow model, integrating prior information for improved results. Other methods \cite{xu2022sinnerf, deng2022depth, roessle2022dense, zhang2021ners, yang2023freenerf, kwak2023geconerf, wang2023sparsenerf} enhance reconstruction by combining different modalities like geometry, depth, and frequency content. CaesarNeRF \cite{zhu2023caesarnerf} utilize global average pooling to encode semantic information. However, it applies the same semantics to all rays, which can lead to conflicts, requiring additional calibration. LVSM \cite{jin2024lvsmlargeviewsynthesis} introduces a scalable, transformer-based model for view synthesis from sparse inputs that relies entirely on data-driven methods, bypassing traditional 3D biases. 

\paragraph{Generalizable 3D Gaussian Splatting.} Recent developments in generalizable Gaussian splatting methods have markedly improved the efficiency of novel view synthesis. These methods bypass the need for per-scene optimization by directly regressing Gaussian parameters via a feed-forward architecture. For instance, PixelSplat \cite{charatan2024pixelsplat} utilizes an epipolar transformer to resolve scale ambiguity and encode features, though its design is limited to image pair inputs. GPS-Gaussian \cite{zheng2024gps} extends this approach to stereo matching, incorporating epipolar rectification, disparity estimation, and feature encoding. However, its reliance on ground-truth depth maps restricts its applicability. Splatter Image \cite{szymanowicz2024splatter} proposes a single-view reconstruction approach based on Gaussian splatting, though its focus on object-centric reconstruction constrains its generalization to unseen scenes. MVSGaussian \cite{liu2024mvsgaussian} combines multi-view stereo with a generalizable 3D Gaussian representation for real-time novel view synthesis. However, its single-sample-per-ray strategy necessitates highly accurate depth estimations, hindering its performance in complex scenes.

\section{Methodology}
Given a set of posed source views of an unknown scene, our goal is to synthesize novel views from arbitrary positions and directions. The proposed GoLF-NRT framework consists of two main stages: 1) constructing a coarse scene representation using global context, and 2) refining this representation by incorporating local geometry through epipolar feature aggregation. As shown in \cref{fig:pipeline}, we begin by extracting multi-scale features from source views. Coarse features, with larger receptive fields, are aggregated to capture global context, forming a preliminary target ray representation. This initial representation is then refined by aggregating fine features with smaller receptive fields along epipolar lines from all source views, resulting in a geometry-aware target ray representation. Finally, a multi-layer perceptron (MLP) decoder predicts the target ray's color.

\subsection{Multi-Scale Feature Extraction}
\label{sec:encoder}
Given $N$ source views $\{ \boldsymbol{I}_i \}_{i=1}^N $ with camera parameters $\{ \boldsymbol{P}_i \}_{i=1}^N $, we employ a feature pyramid network (FPN) \cite{2017Feature} with shared weights to extract multi-scale features at 1/4, 1/8 and 1/16 resolutions. Unlike previous methods that operate on a single scale \cite{varma2022attention, zhu2023caesarnerf, min2024entangled}, our approach leverages features across multiple scales, thereby improving the representational capability of the model. Specifically, we use the 1/8-resolution features for global context extraction, while the 1/4-resolution features for capturing local geometric information.

\begin{figure}[t]
  \centering
  \includegraphics[width=0.95\linewidth]{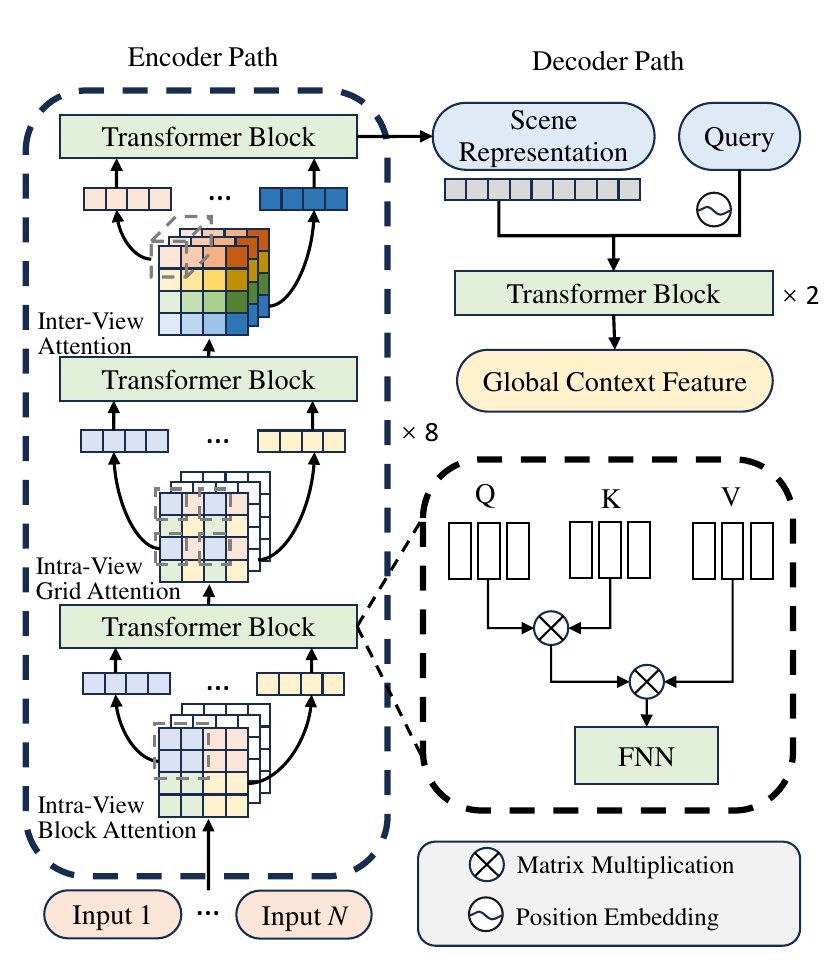}
   \caption{Network architecture of Global Context Feature Extraction Module. The 3D transformer, with near-linear complexity, encodes the scene into a scene representation, while the decoder generates the global context feature for each target ray.}
   \label{fig:global}
\end{figure}

\subsection{Global Context Feature Extraction}
Both cost volume-based \cite{johari2022geonerf} and attention mechanism-based \cite{min2024entangled, suhail2022generalizable} novel view synthesis methods focus heavily on local geometric reasoning for scene generalization but often lack a global perceptual framework, limiting their performance, particularly with a limited number of input views. This limitation arises from their reliance on features along epipolar lines, overlooking rich contextual information from other areas of the source views.

To address this, we propose incorporating features from all source views to create a global context feature for each ray. As described in \cref{sec:encoder}, the encoder aggregates 1/8 resolution features from all source views, generating an implicit scene representation \cite{sajjadi2022scene}, denoted as $\boldsymbol{Z}_g$. However, this approach is computationally intensive and impractical for high-resolution rendering. To address this, we propose a 3D transformer incorporating sparse attention, as illustrated in \cref{fig:global}.

\paragraph{Near-Linear Complexity 3D Transformer.} Directly applying transformers to multi-view, high-resolution images is computationally expensive. To mitigate this, we adopt the sparse attention mechanism from MaxViT \cite{2022MaxViT} and design a 3D transformer with near-linear computational complexity. Full-scale attention is decomposed into three sparse forms: intra-view block attention, intra-view grid attention, and inter-view attention. Intra-view block attention partitions each feature map of shape $H \times W \times C$ into non-overlapping windows of size $P \times P$, reshaping it into a tensor of shape $(\frac{H}{P} \times \frac{W}{P}, P \times P, C)$, and applies self-attention within each window. Similarly, intra-view grid attention divides the tensor into a shape $(G \times G, \frac{H}{G} \times \frac{W}{G}, C)$ using a fixed $G \times G$ uniform grid, resulting in windows with an adaptive size $\frac{H}{G} \times \frac{W}{G}$, and applies self-attention along the grid axis $G \times G$. Finally, inter-view attention gathers feature maps from the $N$ source views into a tensor of shape $(N, H \times W, C)$, and applies self-attention along the view axis $N$. By stacking these attention mechanisms sequentially, we capture both local and global interactions within each block. Notably, both intra-view block attention and intra-view grid attention operate with linear complexity relative to the input size, and the number of views $N$ is typically small. As a result, this design achieves efficient global interaction without incurring the quadratic computational costs of full-scale attention.

To generate the global context feature $\boldsymbol{F}_g$ for each target ray, we use a standard decoder transformer \cite{vaswani2017attention}. This decoder selectively attends to relevant features within the scene representation, providing a preliminary representation of the target ray. For a target ray $\boldsymbol{r}$ with specified origin and direction, the global context feature is computed by querying the decoder transformer $\mathcal{D}$ as follows:
\begin{equation}
    \boldsymbol{F}_g = \mathcal{D} \left( \boldsymbol{r} | \boldsymbol{Z}_g \right). 
\end{equation}
In this mechanism, the target ray serves as the query for multi-head attention, while the keys and values are computed from the scene representation $\boldsymbol{Z}_g$.

Through this interaction, the model selectively extracts the most relevant global context features for the target ray. This offers a comprehensive understanding of the entire scene, allowing the model to more effectively identify crucial changes in scene attributes, even when faced with diverse scenes or few input views.

\subsection{Local Geometric Feature Extraction}

\begin{figure*}[t]
  \centering
  \includegraphics[width=\linewidth]{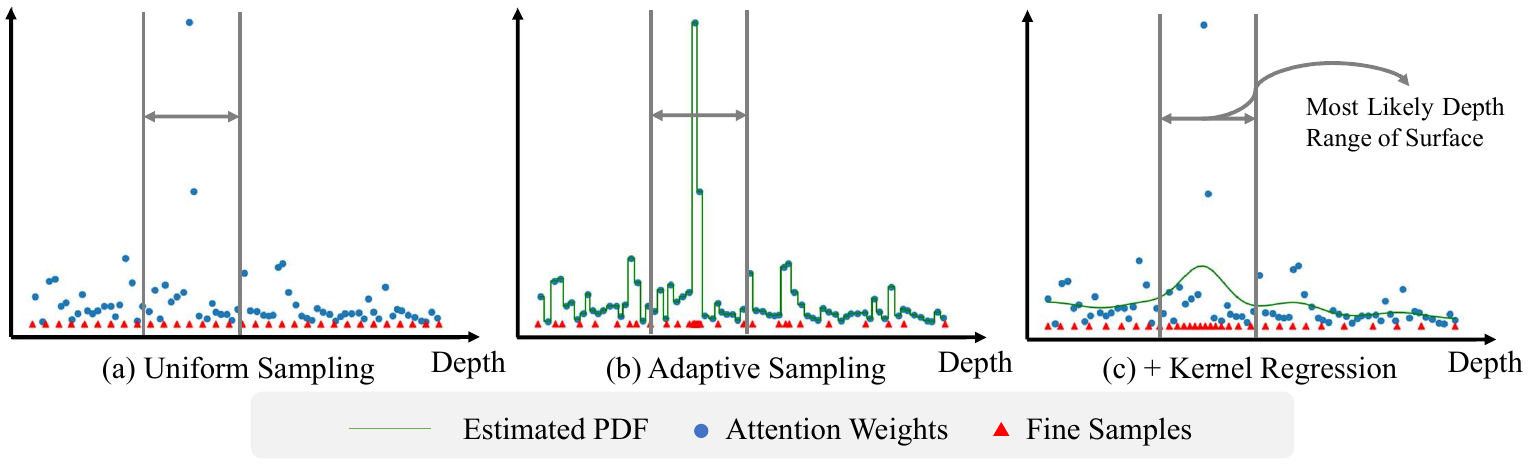}
  \caption{Illustration of how adaptive sampling with kernel regression enhances local geometric perception. (a) Uniform sampling, lacking surface depth constraints, distributes points broadly and randomly. (b) Adaptive sampling based solely on attention weights can lead to disorderly sampling due to the fundamental differences between feature similarity and the PDF used in volume rendering. (c) Our adaptive sampling with kernel regression bridges this gap, focusing samples within the most likely surface depth range, thereby improving subsequent image synthesis.}
  \label{fig:sample}
\end{figure*}

This module constructs a coordinate-aligned feature volume from the 1/4-resolution features $\{ \boldsymbol{F}_i \}_{i=1}^N$ for the target ray, aiming to capture local geometry. We utilize GNT \cite{varma2022attention}, a fully attention-based generalizable NeRF method, as our backbone. GNT aggregates features along epipolar lines in source views through two transformer-based stages. In the first stage, the view transformer predicts coordinate-aligned features for each point by aggregating information from the epipolar lines in the source views. This stage is designed to establish correspondences between queried points and the source views, providing a likelihood score that indicates the correlation of a pixel on the source view to the same point in 3D space without occlusion. Consequently, the output of the view transformer yields local geometric features of the pixels. In the second stage, the ray transformer aggregates these point-wise features along the ray to compute the ray color. However, while GNT performs well with a larger number of input views, it may encounter challenges when fewer views are available, primarily due to inconsistencies introduced by occlusion or reflections, which can compromise feature aggregation by the view transformer. To address these limitations, we propose two key improvements: global context-guided aggregation and adaptive sampling.

\paragraph{Global Context-Guided Aggregation.} We enhance the view transformer by incorporating the global context feature $\boldsymbol{F}_g$ as the initial query, replacing the element-wise max pooling used in previous methods \cite{varma2022attention}:
\begin{equation}
    \boldsymbol{F}_l = \textit{View-Transformer}(\boldsymbol{F}_g | \{\boldsymbol{F}_i, \boldsymbol{P}_i\}_{i=1}^{N}), 
\end{equation}
where $\boldsymbol{F}_l$ denotes the local geometric feature. By leveraging the global context feature, which containing the most relevant information about the target ray in the scene representation $\boldsymbol{Z}_g$, the view transformer can more accurately assess contributions from various source views, resulting in more precise local geometric features $\boldsymbol{F}_l$. These local features are subsequently concatenated with the global context feature to create a global-local embedding:
\begin{equation}
    \boldsymbol{F}_{g \text{-} l} = \boldsymbol{F}_g \oplus \boldsymbol{F}_l, 
\end{equation}
which is then aggregated by the ray transformer:
\begin{equation}
    \boldsymbol{F}_r = \textit{Ray-Transformer}(\boldsymbol{F}_{g \text{-} l}). 
\end{equation}
After iterating through 8 view and ray transformer blocks, we obtain the target ray representation $\boldsymbol{\hat{F}}_r$. These improvements enhance the robustness of our model in handling challenging scenarios with limited input views, significantly advancing the accuracy of local geometric feature extraction.

\begin{table*}[t]
\small
\centering
\begin{tabular}{ccccccccccc}
\hline
\multirow{2}{*}{Input} & \multirow{2}{*}{Methods} & \multicolumn{3}{c}{LLFF} & \multicolumn{3}{c}{Blender} & \multicolumn{3}{c}{Shiny} \\
\cmidrule(r){3-5} \cmidrule(r){6-8} \cmidrule(r){9-11}
 & & PSNR $\uparrow$ & SSIM $\uparrow$ & LPIPS $\downarrow$ & PSNR $\uparrow$ & SSIM $\uparrow$ & LPIPS $\downarrow$ & PSNR $\uparrow$ & SSIM $\uparrow$ & LPIPS $\downarrow$ \\
\hline
\multirow{7}{*}{3-view} 
& IBRNet \cite{wang2021ibrnet} & 23.00 & 0.752 & 0.262 & 22.44 & 0.874 & 0.195 & 21.96 & 0.710 & 0.281 \\
 & MVSNeRF \cite{chen2021mvsnerf} & 19.84 & 0.729 & 0.314 & 23.62 & 0.897 & 0.176 & 18.55 & 0.645 & 0.343 \\
 & MatchNeRF \cite{chen2023explicit} & 22.30 & 0.731 & 0.234 & 23.20 & 0.897 & 0.164 & 20.77 & 0.672 & 0.249 \\
 & MVSGaussian \cite{liu2024mvsgaussian} & 24.07 & 0.857 & 0.164 & \underline{25.54} & \pmb{0.944} & \pmb{0.073} & 20.49 & 0.661 & 0.254 \\
 & GNT \cite{varma2022attention} & 23.28 & 0.768 & 0.230 & \pmb{25.80} & 0.905 & 0.104 & 22.47 & 0.720 & 0.247 \\
 & EVE-NeRF \cite{min2024entangled} & 22.79 & 0.754 & 0.226 & 23.43 & 0.903 & 0.132 & \underline{24.11} & \pmb{0.781} & \pmb{0.204} \\
 & CaesarNeRF \cite{zhu2023caesarnerf} & \underline{23.45} & \underline{0.794} & \underline{0.176} & 23.56 & 0.908 & 0.131 &22.74 & 0.723 & 0.241 \\
 & GoLF-NRT & \pmb{24.20} & \pmb{0.821} & \pmb{0.148} & 24.30 & \underline{0.916} & \underline{0.097} & \pmb{25.06} & \underline{0.765} & \underline{0.207} \\
\hline
\multirow{6}{*}{2-view} 
 & MVSNeRF \cite{chen2021mvsnerf} & 19.15 & 0.704 & 0.336 & 20.56 & 0.856 & 0.243 & 17.25 & 0.577 & 0.416 \\
 & MatchNeRF \cite{chen2023explicit} & 21.08 & 0.689 & 0.272 & 20.57 & 0.864 & 0.200 & 20.28 & 0.636 & \underline{0.278} \\
 & pixelSplat \cite{charatan2024pixelsplat} & \pmb{22.99} & \pmb{0.810} & \underline{0.190} & 15.77 & 0.755 & 0.314 & - & - & - \\
 & GNT \cite{varma2022attention} & 20.94 & 0.687 & 0.301 & \pmb{23.47} & 0.877 & \underline{0.151} & 20.42 & 0.617 & 0.327 \\
 & EVE-NeRF \cite{min2024entangled} & 19.95 & 0.607 & 0.340 & 21.62 & 0.867 & 0.180 & 20.89 & \underline{0.667} & 0.321 \\
 & CaesarNeRF \cite{zhu2023caesarnerf} & 21.94 & 0.736 & 0.224 & 22.07 & \underline{0.879} & 0.176 & \underline{21.47} & 0.652 & 0.293 \\
 & GoLF-NRT & \underline{22.48} & \underline{0.765} & \pmb{0.193} & \underline{22.34} & \pmb{0.886} & \pmb{0.143} & \pmb{23.03} & \pmb{0.692} & \pmb{0.261} \\
\hline
\multirow{3}{*}{1-view} 
& GNT \cite{varma2022attention} & 16.60 & 0.491 & 0.514 & 15.53 & 0.634 & 0.371 & 15.99 & 0.400 & 0.548 \\
 & CaesarNeRF \cite{zhu2023caesarnerf} & \pmb{18.31} & \underline{0.521} & \underline{0.435} & \underline{15.76} & \underline{0.698} & \underline{0.366} & \underline{17.57} & \underline{0.472} & \underline{0.467} \\
 & GoLF-NRT & \pmb{18.31} & \pmb{0.527} & \pmb{0.410} & \pmb{16.34} & \pmb{0.706} & \pmb{0.309} & \pmb{18.96} & \pmb{0.513} & \pmb{0.393} \\
\hline
\end{tabular}
\caption{Quantitative comparison of state-of-the-art view synthesis methods on LLFF, Blender, and Shiny datasets with 1-3 input views. Best results are in bold, second-best are underlined.}
\label{tab:few-shot}
\end{table*}

\paragraph{Adaptive Sampling.} After the view transformer obtains point-wise features along the ray, the ray transformer aggregates them to form the ray color. The weighted aggregation can be regarded as an approximation of volume rendering \cite{kajiya1984ray}, where the weights are learned through the attention mechanism. This insight inspires us to explore incorporating established volume rendering techniques to enhance modeling efficiency, such as hierarchical volume sampling, which prioritizes regions with significant contributions to the final rendering. We propose using the attention weights in the ray transformer as accumulation weights for volume rendering, similar to a probability density function (PDF). However, attention weights, derived from feature similarity, differ fundamentally from traditional PDFs used in volume rendering. As shown in \cref{fig:sample}(b), while attention weights provide valuable cues, the estimated PDF can lead to disorderly sampling.

To bridge the gap between attention weights and the PDF, we introduce a kernel regression method \cite{2005Image,2006Robust,2007Kernel} that transforms attention weights into a smooth, regularized PDF. Given $N_c$ uniformly sampled depths $\{ d_i \}_{i=1}^{N_c}$ along the ray with corresponding attention weights $\{ w_i \}_{i=1}^{N_c}$, the PDF at depth $d$ is calculated as follows:
\begin{equation}
     p(d) = \sum_{i=1}^{N_c} w_i * \frac{K(d, d_i)}{\sum_{i=1}^{N_c} K(d, d_i)}, 
\label{eq:pdf}
\end{equation}
where $K(d, d_i)$ is the Gaussian kernel:
\begin{equation}
    K(d, d_i) = \frac{1}{h \sqrt{2\pi}} e^{-0.5(\frac{d - d_i}{h})^2}, 
\end{equation}
and $h$ is the bandwidth parameter. This method not only smooths the noise but also ensures that the resulting sampling distributions aligns with realistic, physically plausible shapes—either unimodal for opaque objects or multimodal for translucent materials, as depicted in \cref{fig:sample}(C).

In the uniform sampling stage, $N_c$ features are uniformly sampled to estimate the PDF using \cref{eq:pdf}. Subsequently, $N_f$ refined feature locations are sampled from the PDF using inverse transform sampling, biasing the samples towards the object's surface. In the adaptive sampling stage, all $N_c + N_f$ samples are used. Both stages employ the same transformer.

\subsection{MLP Decoder}
Given the final feature of the target ray $\boldsymbol{\hat{F}}_r$, the ray color is then obtained through a multi-layer perceptron (MLP) decoder:
\begin{equation}
    \boldsymbol{\hat{c}} = \mathrm{MLP}(\boldsymbol{\hat{F}}).
\end{equation}

\section{Experiments}
\subsection{Settings}
\paragraph{Dataset.} Following \cite{wang2021ibrnet, varma2022attention}, we constructed our training data from both synthetic and real sources, including 1023 models from Google Scanned Object \cite{downs2022google}, RealEstate10K \cite{zhou2018stereo}, Spaces scenes \cite{flynn2019deepview}, and 102 real scenes captured with handheld cellphones \cite{mildenhall2019local, wang2021ibrnet}. For evaluation, we used the LLFF \cite{mildenhall2019local}, Blender \cite{mildenhall2021nerf}, and Shiny \cite{wizadwongsa2021nex} datasets. Notably, LLFF validation scenes were excluded from training, and Blender and Shiny datasets introduce significant variations in view distribution and scene content compared to the training dataset. The quality of the synthesized novel views is measured using the PSNR, SSIM \cite{wang2004image}, and LPIPS \cite{zhang2018unreasonable} metrics.

\paragraph{Implementation Details.} GoLF-NRT is trained using the Mean Square Error (MSE) loss commonly used in NeRF \cite{mildenhall2021nerf}. We optimized the model end-to-end using the Adam optimizer, setting an initial learning rate of $10^{-3}$ for the multi-scale feature extraction module and $5 \times 10^{-4}$ for other modules, with exponential decay. Training ran for 250k iterations with 682 rays sampled per iteration. During uniform sampling, we used 128 samples per ray, with an additional 64 samples during adaptive sampling. Following \cite{varma2022attention}, we construct source-target view pairs by selecting a target view and forming a pool of $k \times N$ proximate views, where $k \in (1, 3)$ and $N \in (8, 12)$. During training, $k$ and $N$ are uniformly sampled from these intervals, while $N$ is fixed at 10 for evaluation to ensure consistency.



\begin{figure*}[t]
 \centering
\subfloat[3 input views]{\includegraphics[width=0.95\linewidth]{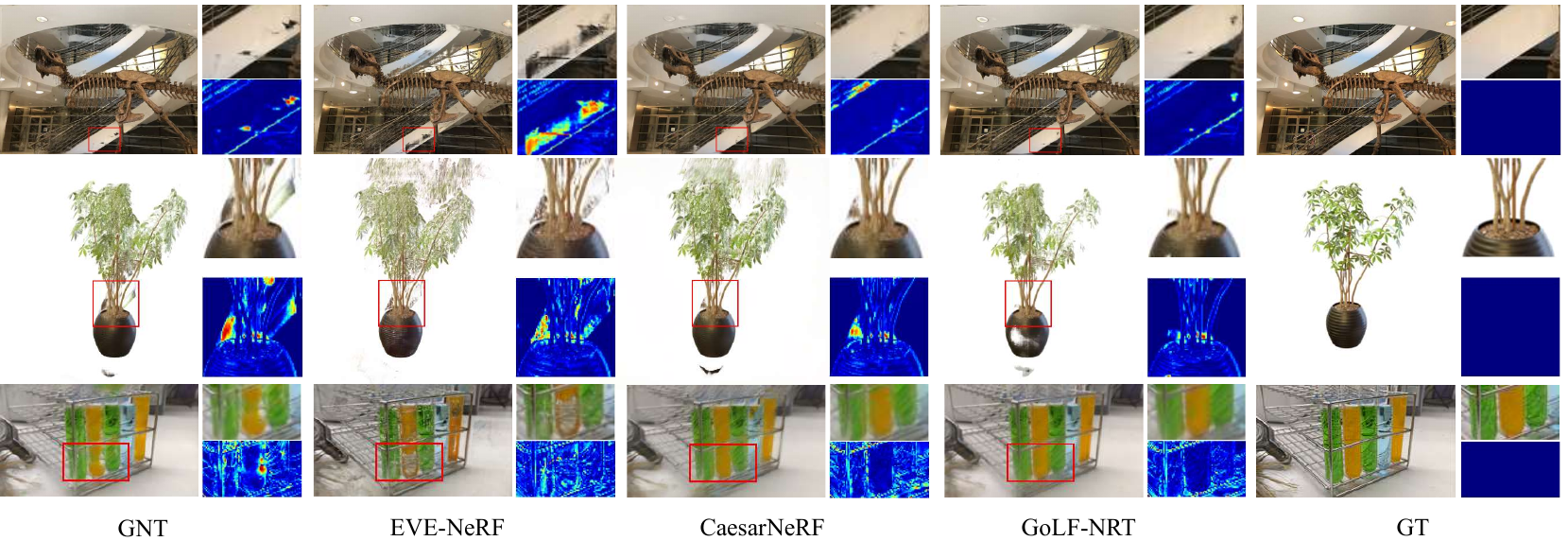}} \\
\subfloat[2 input views]{\includegraphics[width=0.95\linewidth]{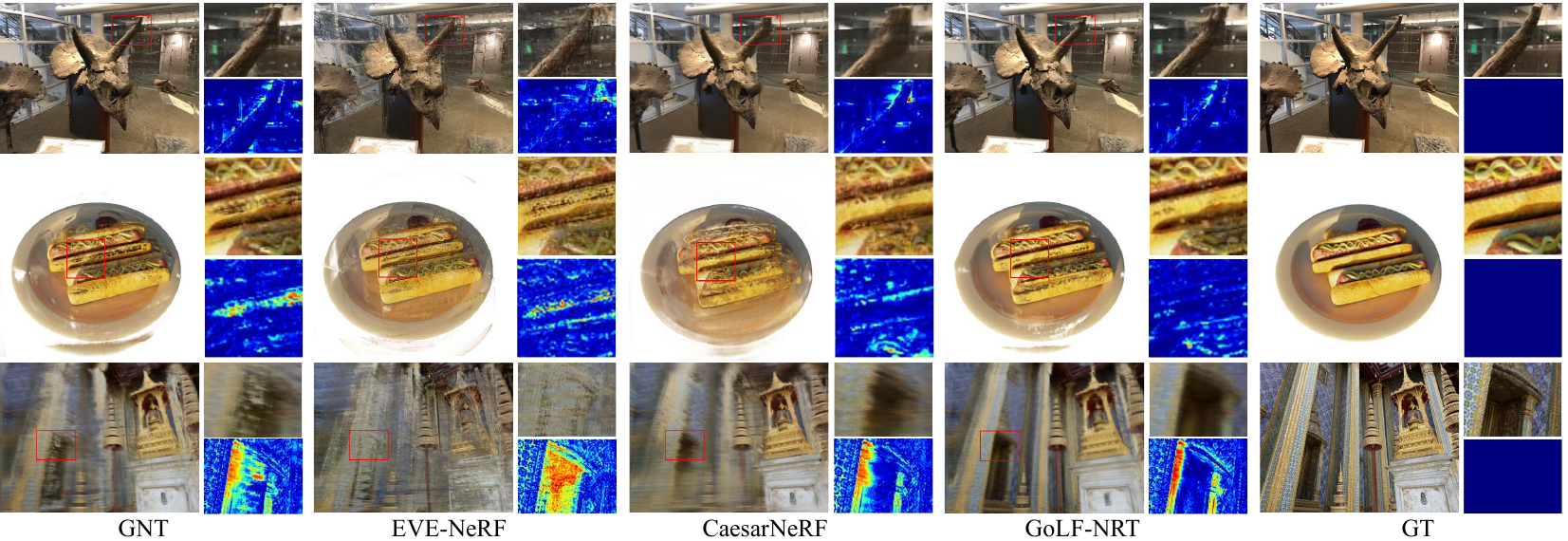}}
\caption{Qualitative comparison of GoLF-NRT with GNT, EVE-NeRF, and CaesarNeRF using 3 and 2 input views: (a) Trex scene (LLFF), Ficus scene (Blender), and Lab scene (Shiny); (b) Horns scene (LLFF), Hotdog scene (Blender), and Crest scene (Shiny). Each triplet shows the reconstructed image (left), a zoomed-in view (upper right), and its corresponding error map (lower right).}
\label{fig:few-shot}
\end{figure*}

\subsection{Comparative Studies}
We compare GoLF-NRT with several state-of-the-art generalizable NeRF methods, including PixelNeRF \cite{yu2021pixelnerf}, IBRNet \cite{wang2021ibrnet}, MVSNeRF \cite{chen2021mvsnerf}, MatchNeRF \cite{chen2023explicit}, NeuRay \cite{liu2022neural}, GPNR \cite{suhail2022generalizable}, GNT \cite{varma2022attention}, CaesarNeRF \cite{zhu2023caesarnerf}, and EVE-NeRF \cite{min2024entangled}. We conducted experiments under both few-shot and many-shot configurations to provide a comprehensive comparison with existing methods, showcasing the broad applicability and superiority of GoLF-NRT.

As shown in \cref{tab:few-shot}, GoLF-NRT outperforms other methods across most metrics with 1 to 3 input views on LLFF, Blender, and Shiny. While all methods degrade with fewer views, our approach maintains high-quality results with minimal loss. Notably, with 3 input views on the LLFF dataset, GoLF-NRT surpasses CaesarNeRF \cite{zhu2023caesarnerf} by 0.75 dB in PSNR, 0.027 in SSIM, and reduces LPIPS by 0.028. It's also important to note that multi-view matching becomes irrelevant when only a single image is available, explaining why our results significantly outperform GNT \cite{varma2022attention}. While CaesarNeRF also leverages semantic information for few-shot novel view synthesis, it applies identical semantics to all rays, which can lead to conflicts, even with additional calibration. In contrast, our approach selectively extracts relevant global information for each ray, using it to guide local information extraction, thereby supporting accurate ray synthesis even on complex datasets such as Shiny dataset. With 1 input view on the Shiny dataset, GoLF-NRT improves over CaesarNeRF by 1.39 dB in PSNR, 0.041 in SSIM, and reduces LPIPS by 0.074.

\cref{fig:few-shot} presents qualitative results for configurations with 3 and 2 input views. It is evident that previous methods relying solely on local features for multi-view matching encounter difficulties in regions with edges, depth discontinuities, or reflections, often resulting in mismatches. By incorporating global features within the view transformer, our approach achieves a more holistic scene understanding, leading to enhanced consistency in matching across views.

\begin{figure*}[!htbp]
  \centering
  \includegraphics[width=0.95\linewidth]{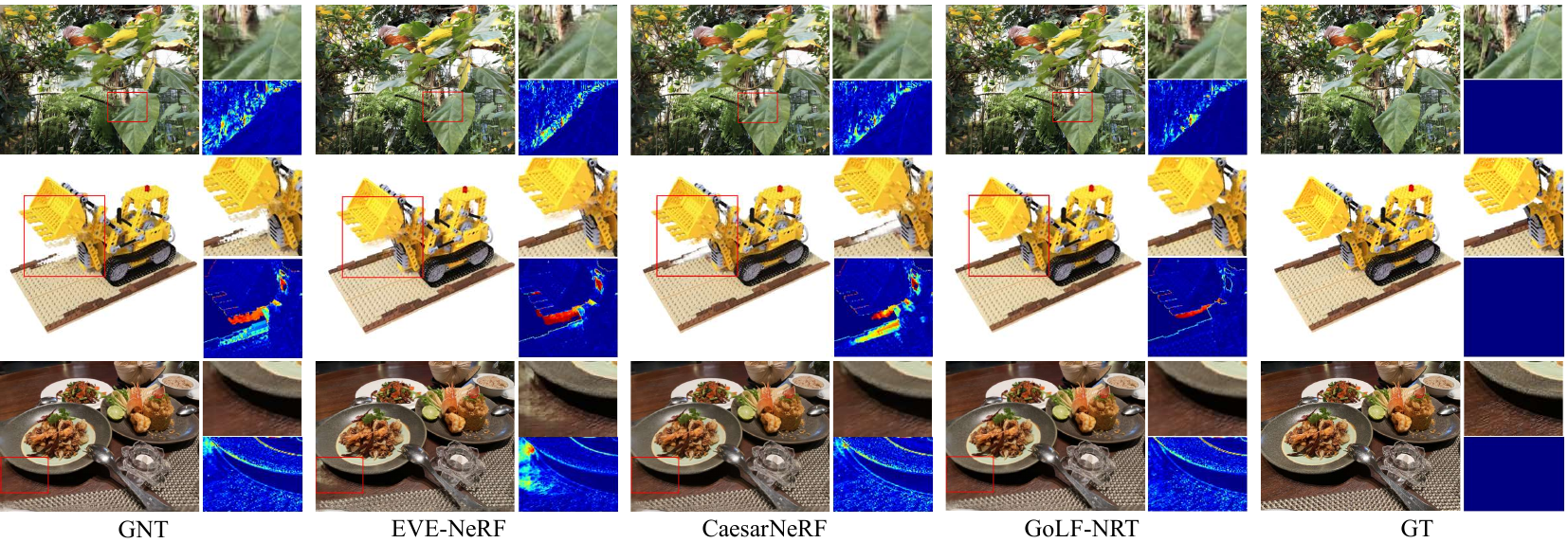}
  \caption{Qualitative comparison of GoLF-NRT with GNT, EVE-NeRF, and CaesarNeRF using 10 input views. Rows correspond to the Leaves scene (LLFF), Lego scene (Blender), and Food scene (Shiny). Each triplet includes a reconstructed image (left), a zoomed-in view (upper right), and its error map (lower right).}
  \label{fig:10-views}
\end{figure*}

We also demonstrate generalizable results under many-shot conditions (10-views), as shown in \cref{tab:many-shot}. Our method achieved the best or second-best results across all metrics. \cref{fig:10-views} provides a qualitative comparison, confirming its effectiveness in both few-shot and many-shot scenarios.

\begin{table*}[!htbp]
\small
\centering
\begin{tabular}{cccccccccc}
\hline
\multirow{2}{*}{Methods} & \multicolumn{3}{c}{LLFF} & \multicolumn{3}{c}{Blender} & \multicolumn{3}{c}{Shiny} \\
\cmidrule(r){2-4} \cmidrule(r){5-7} \cmidrule(r){8-10}
 & PSNR $\uparrow$ & SSIM $\uparrow$ & LPIPS $\downarrow$ & PSNR $\uparrow$ & SSIM $\uparrow$ & LPIPS $\downarrow$ & PSNR $\uparrow$ & SSIM $\uparrow$ & LPIPS $\downarrow$ \\
\hline
PixelNeRF \cite{yu2021pixelnerf} & 18.66 & 0.588 & 0.463 & 22.65 & 0.808 & 0.202 & - & - & -  \\
IBRNet \cite{wang2021ibrnet} & 25.17 & 0.813 & 0.200 & 26.73 & 0.908 & 0.101 & 23.60 & 0.785 & 0.180 \\
NeuRay \cite{liu2022neural} & 25.35 & 0.818 & 0.198 & 26.48 & 0.944 & 0.091 & 25.72 & 0.880 & 0.175 \\
GPNR \cite{suhail2022generalizable} & 25.72 & \pmb{0.880} & 0.175 & \underline{28.29} & 0.927 & 0.080 & 24.12 & 0.860 & 0.170 \\
GNT \cite{varma2022attention} & 25.53 & 0.836 & 0.178 & 26.01 & 0.925 & 0.088 & 26.56 & 0.852 & 0.132 \\
CaesarNeRF \cite{zhu2023caesarnerf} &  25.16 & 0.851 & \underline{0.138} & 26.75 & 0.915 & 0.083 & 26.66 & 0.849 & 0.146 \\
EVE-NeRF \cite{min2024entangled} & \pmb{27.16} & 0.869 & 0.141 & 27.03 & \underline{0.952} & \underline{0.072} & \underline{27.41} & \underline{0.883} & \underline{0.113} \\
\hline
GoLF-NRT & \underline{26.42} & \underline{0.879} & \pmb{0.109} & \pmb{29.74} & \pmb{0.963} & \pmb{0.060} & \pmb{28.01} & \pmb{0.885} & \pmb{0.112} \\
\hline
\end{tabular}
\caption{Quantitative comparison with state-of-the-art view synthesis methods on LLFF, Blender and Shiny datasets with 10 input views. Best results are in bold, second-best are underlined.}
\label{tab:many-shot}
\end{table*}

\subsection{Ablations and Analysis}
\begin{table}[!htbp]
\small
\centering
\begin{tabular}{cccc}
\hline
Model & PSNR$\uparrow$ & SSIM$\uparrow$ & LPIPS$\downarrow$ \\
\hline
only local geometry & 23.28 & 0.768 & 0.230 \\
+ global context & 23.55 & 0.791 & 0.170 \\ 
+ global context as query & 23.72 & 0.804 & 0.159 \\
+ adaptive sampling & 23.88 & 0.806 & 0.157 \\
+ kernel regression & \pmb{24.20} & \pmb{0.821} & \pmb{0.148} \\
\hline
\end{tabular}
\caption{Ablations. We trained model variants using the previously described setup and evaluated them on the LLFF dataset with 3 input views.}
\label{tab:ablation}
\end{table}

\begin{figure}[!htbp]
  \centering
  \includegraphics[width=0.95\linewidth]{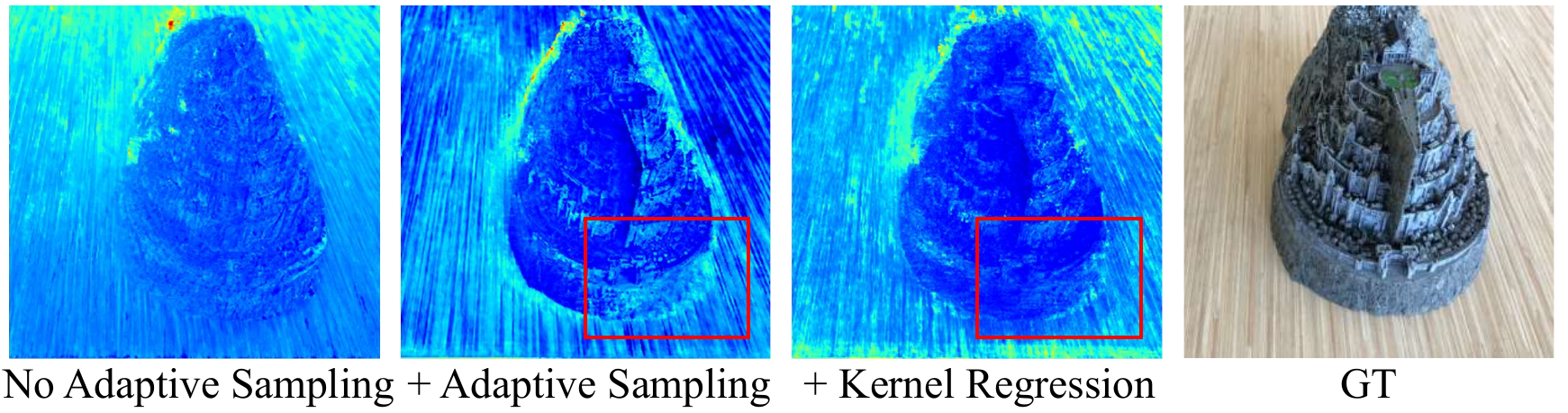}
  \caption{The depth maps from three sampling strategies show that adaptive sampling with kernel regression significantly improves accuracy, especially at boundaries with depth discontinuities.}
  \label{fig:depth}
\end{figure}

\begin{figure}[!htbp]
    \centering
    \includegraphics[width=0.48\textwidth]{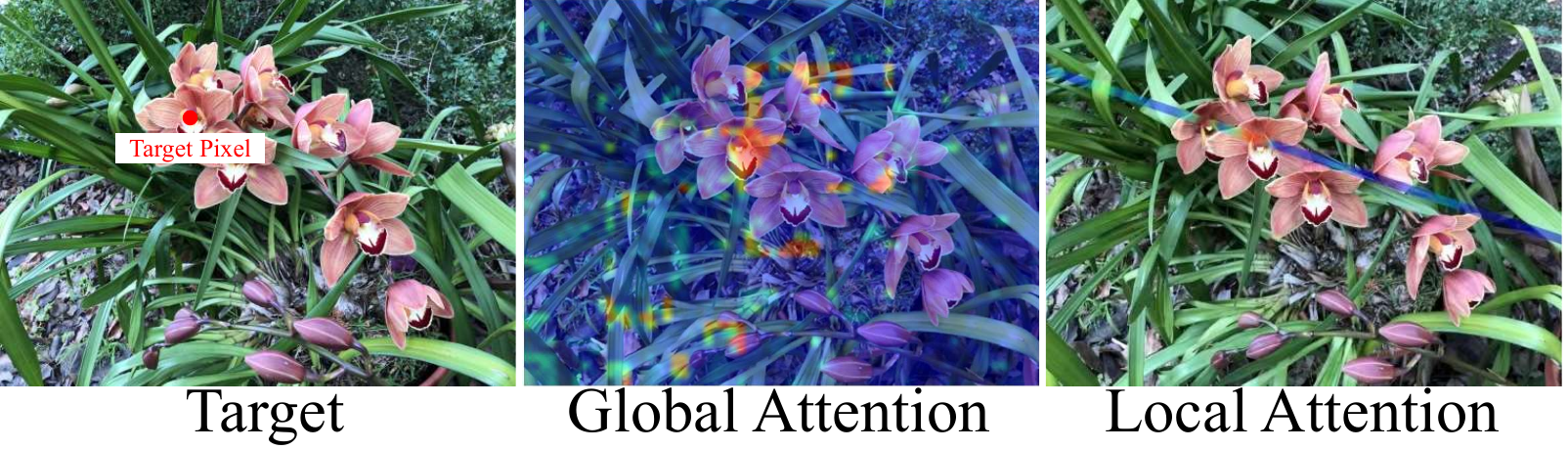}
    \caption{Visualization of global and local attentions.} 
    \label{fig:attn}
\end{figure}

We conduct a series of ablation studies to evaluate the effectiveness of our design. \cref{tab:ablation} shows that our method, which integrates both global and local features, improves PSNR by 0.44 dB, underscoring the importance of global context in achieving accurate novel view synthesis. \cref{fig:attn} illustrates that our design effectively focuses on important regions. Adaptive sampling using kernel regression on attention weights reduces noise from local similarities and produces a more targeted sampling distribution, resulting in a 0.32 dB increase in PSNR. \cref{fig:depth} shows that this significantly enhances depth recovery at object boundaries. Additionally, \cref{tab:flops} shows that full-scale attention is computationally impractical, while our 3D transformer reduces costs, enabling high-resolution processing.

\begin{table}[!htbp]
\small
\centering
\begin{tabular}{ccc}
\hline
 & FLOPs (G) & Parameters (M) \\
\hline
Full Attention & 11.67 & 0.37 \\
Our 3D Transformer & 1.88 & 0.33 \\
\hline
\end{tabular}
\caption{Comparison of efficiency with full-scale attention.}
\label{tab:flops}
\end{table}
\section{Conclusion}
In this paper, we introduced GoLF-NRT, a novel neural rendering transformer that integrates global and local features to enhance view synthesis from few input views. Our approach utilizes a 3D transformer with efficient sparse attention for global context understanding and an adaptive sampling strategy for accurate local geometry perception, significantly improving rendering quality. Extensive experiments demonstrated that GoLF-NRT achieves state-of-the-art performance across various datasets, particularly in scenarios with limited input views. Our method’s ability to handle depth ambiguity and occlusions highlights its robustness and versatility, making it a promising solution for real-world applications in neural rendering.
{
    \small
    \bibliographystyle{ieeenat_fullname}
    \bibliography{main.bib}
}
\clearpage
\maketitleagain{Supplementary Material}

\begin{table*}[t]
\small
\centering
\begin{tabular}{ccccccc}
\hline
Metric & Method & cd & crest & food & giants & lab\\
\hline
\multirow{3}{*}{PSNR $\uparrow$} & GNT \cite{varma2022attention} & 32.39 & 21.79 & \underline{25.68} & 27.95 & \underline{25.01} \\
 & EVE-NeRF \cite{min2024entangled} & \underline{33.73} & \underline{23.59} & 25.61 & \underline{29.24} & 24.87 \\
 & GoLF-NRT & \pmb{34.10} & \pmb{23.72} & \pmb{26.06} & \pmb{29.31} & \pmb{25.85} \\
\hline
\multirow{3}{*}{SSIM $\uparrow$} & GNT \cite{varma2022attention} & 0.959 & 0.692 & 0.853 & 0.908 & 0.850 \\
 & EVE-NeRF \cite{min2024entangled} & \underline{0.971} & \pmb{0.777} & \underline{0.856} & \underline{0.924} & \underline{0.888} \\
 & GoLF-NRT & \pmb{0.979} & \underline{0.768} & \pmb{0.864} & \pmb{0.928} & \pmb{0.895} \\
\hline
\multirow{3}{*}{LPIPS $\downarrow$}  & GNT \cite{varma2022attention} & 0.056 & 0.237 & 0.128 & 0.093 & \underline{0.144} \\
 & EVE-NeRF \cite{min2024entangled} & \underline{0.038} & \pmb{0.169} & \underline{0.127} & \underline{0.085} & 0.146 \\
 & GoLF-NRT & \pmb{0.037} & \underline{0.189} & \pmb{0.118} & \pmb{0.083} & \pmb{0.136} \\
\hline
\end{tabular}
\caption{Per-scene quantitative comparison of state-of-the-art view synthesis methods on Shiny with 10 input views. Best results are in bold, second-best are underlined.}
\label{tab:per-shiny}
\end{table*}

\begin{table*}[t]
\small
\centering
\begin{tabular}{cccccccccc}
\hline
Metric & Method & trex & fern & flower & leaves & room & fortress & horns & orchids \\
\hline
\multirow{5}{*}{PSNR $\uparrow$} & LLFF \cite{mildenhall2019local} & 27.48 & \pmb{28.72} & 20.72 & 21.13 & 24.54 & 21.79 & 23.22 & 18.52 \\
 & NeRF \cite{mildenhall2021nerf} & 26.80 & 25.17 & 27.40 & 20.92 & 32.70 & 31.16 & 27.45 & 20.36 \\
 & NeX \cite{wizadwongsa2021nex} & \underline{28.73} & 25.63 & \pmb{28.90} & 21.96 & 32.32 & 31.67 & 28.46 & 20.42 \\
 & GNT \cite{varma2022attention} & 28.15 & 24.31 & 27.32 & \underline{22.57} & \underline{32.96} & \underline{32.28} & \underline{29.62} & 2\underline{0.67} \\
  & GoLF-NRT & \pmb{28.75} & \underline{25.65} & \underline{28.77} & \pmb{23.18} & \pmb{33.09} & \pmb{32.47} & \pmb{29.66} & \underline{21.27} \\
\hline
\multirow{5}{*}{SSIM $\uparrow$} & LLFF \cite{mildenhall2019local} & 0.857 & 0.753 & 0.844 & 0.697 & 0.932 & 0.872 & 0.840 & 0.588 \\
 & NeRF \cite{mildenhall2021nerf} & 0.880 & 0.792 & 0.827 & 0.690 & 0.948 & 0.881 & 0.828 & 0.641 \\
 & NeX \cite{wizadwongsa2021nex} & \pmb{0.953} & \pmb{0.887} & \pmb{0.933} & 0.832 & \pmb{0.975} & \pmb{0.952} & \underline{0.937} & \underline{0.765} \\
 & GNT \cite{varma2022attention} & 0.936 & 0.846 & 0.893 & \underline{0.852} & 0.963 & 0.934 & 0.935 & 0.752 \\
 & GoLF-NRT & \underline{0.944} & \underline{0.851} & \underline{0.915} & \pmb{0.873} & \underline{0.971} & \underline{0.936} & \pmb{0.941} & \pmb{0.787} \\
\hline
\multirow{5}{*}{LPIPS $\downarrow$} & LLFF \cite{mildenhall2019local} & 0.222 & 0.247 & 0.174 & 0.216 & 0.155 & 0.173 & 0.193 & 0.313 \\
 & NeRF \cite{mildenhall2021nerf} & 0.249 & 0.280 & 0.219 & 0.316 & 0.178 & 0.171 & 0.263 & 0.321 \\
 & NeX \cite{wizadwongsa2021nex} & 0.193 & 0.205 & 0.150 & 0.173 & 0.161 & 0.131 & 0.173 & 0.242 \\
 & GNT \cite{varma2022attention} & \underline{0.080} & \underline{0.116} & \underline{0.092} & \underline{0.109} &\underline{ 0.060} & \underline{0.061} & \underline{0.076} & \underline{0.153} \\
 & GoLF-NRT & \pmb{0.076} & \pmb{0.114} & \pmb{0.068} & \pmb{0.092} & \pmb{0.057} & \pmb{0.058} & \pmb{0.070} & \pmb{0.136} \\
\hline
\end{tabular}
\caption{Per-scene optimization quantitative comparison of state-of-the-art view synthesis methods on LLFF with 10 input views. Best results are in bold, second-best are underlined.}
\label{tab:per-scene}
\end{table*}

\begin{figure*}[!ht]  
\centering
\includegraphics[width=0.95\linewidth]{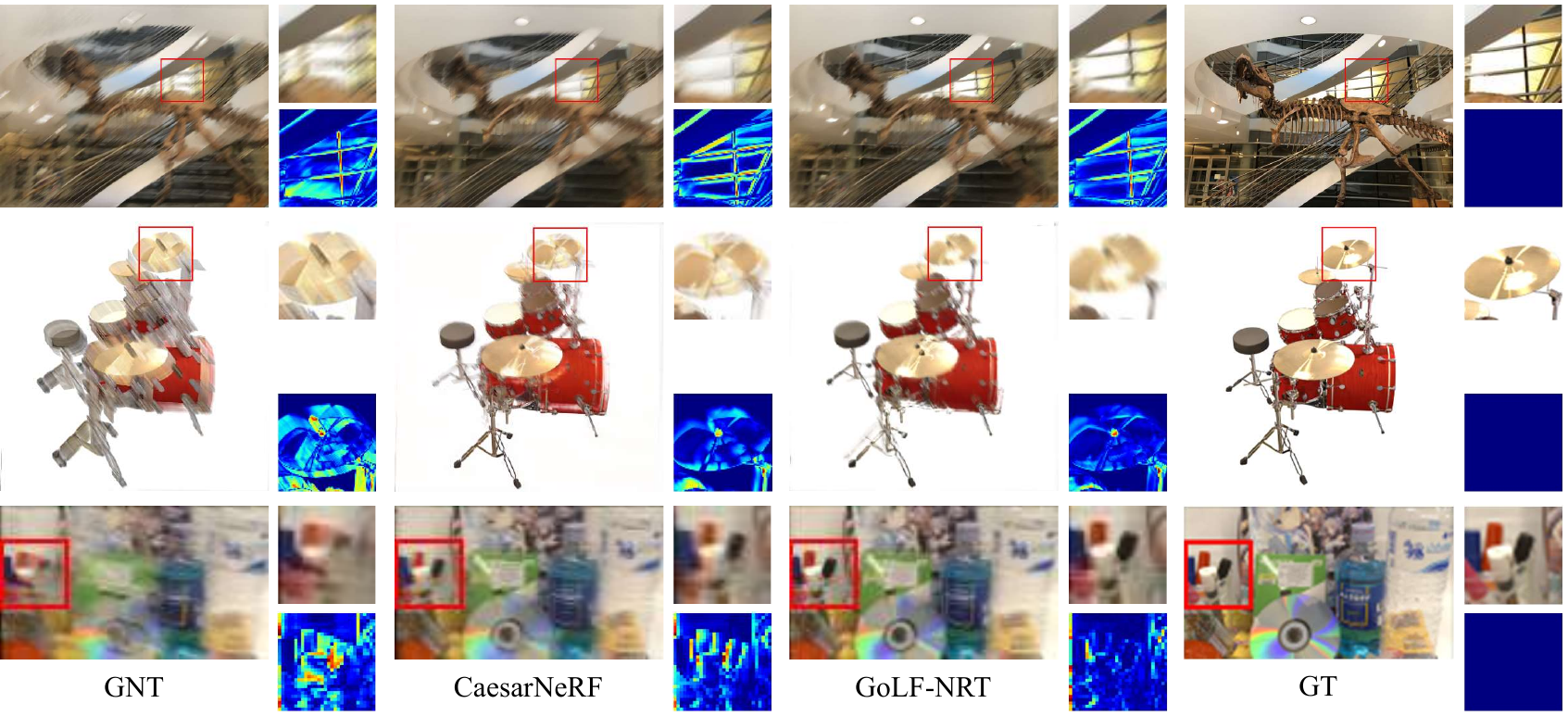} 
\caption{Qualitative comparison of GoLF-NRT with GNT, EVE-NeRF and CaesarNeRF with 1 input views. The first, second, and third rows correspond to the Trex scene from LLFF, the Drums scene from Blender, and the CD scene from Shiny, respectively. Each image triplet includes: the reconstructed image on the left, a zoomed-in view on the upper right, and the error map corresponding to the zoomed-in view on the lower right.}
\label{fig:1view}
\end{figure*}

\begin{figure*}[!ht]  
\centering
\includegraphics[width=0.95\linewidth]{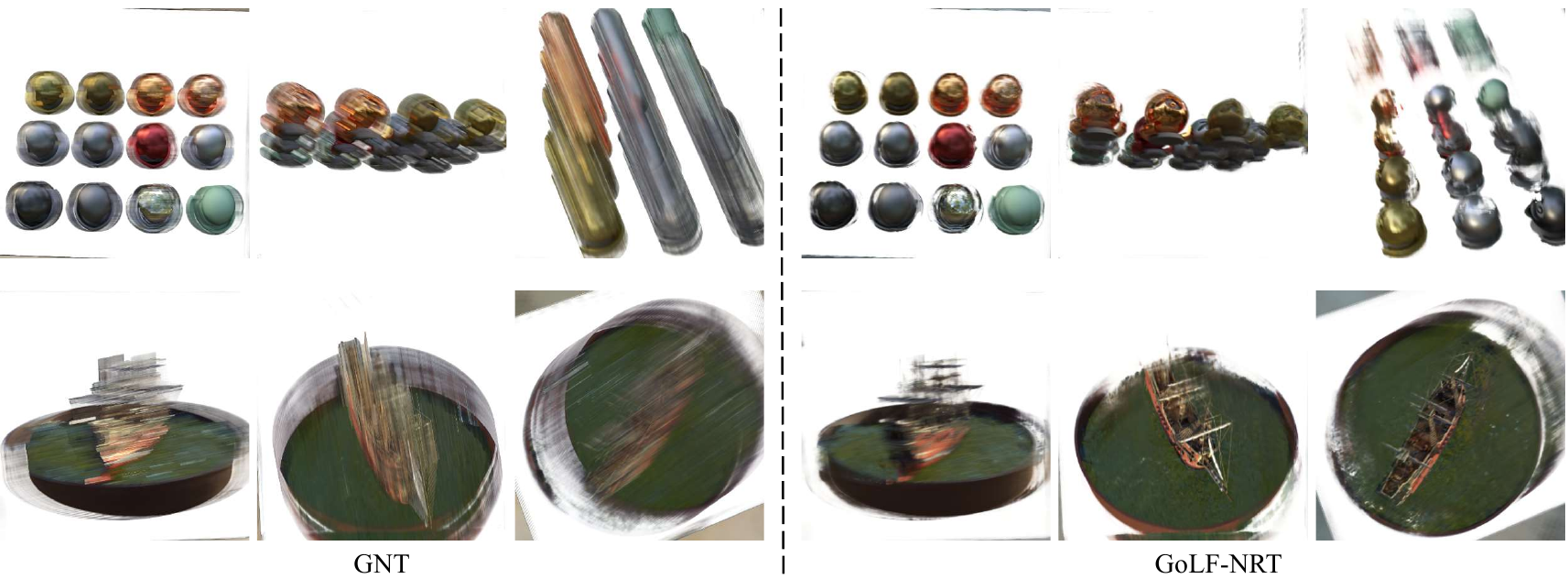} 
\caption{Qualitative comparison between GoLF-NRT and GNT with 1 input views. The first and secondrows correspond to the Materials and Ship scene from Blender, respectively.}
\label{fig:1nerf}
\end{figure*}

\begin{figure*}[!ht]  
\centering
\includegraphics[width=0.95\linewidth]{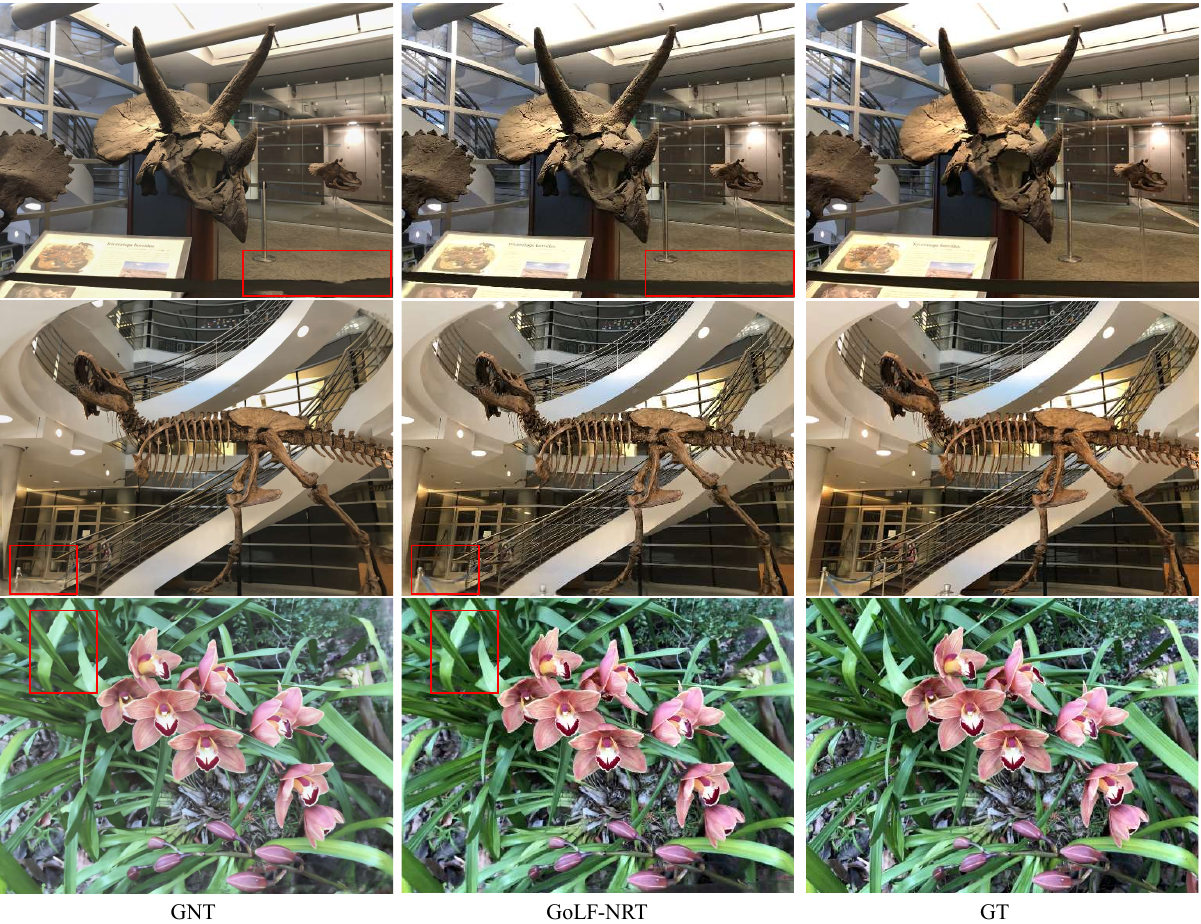} 
\caption{Qualitative comparison between GoLF-NRT and GNT with 10 input views. The first, second, and third rows correspond to the Horns, Trex, and Orchids scene from LLFF, respectively.}
\label{fig:10per}
\end{figure*}

\begin{figure*}[!ht]  
\centering
\includegraphics[width=0.95\linewidth]{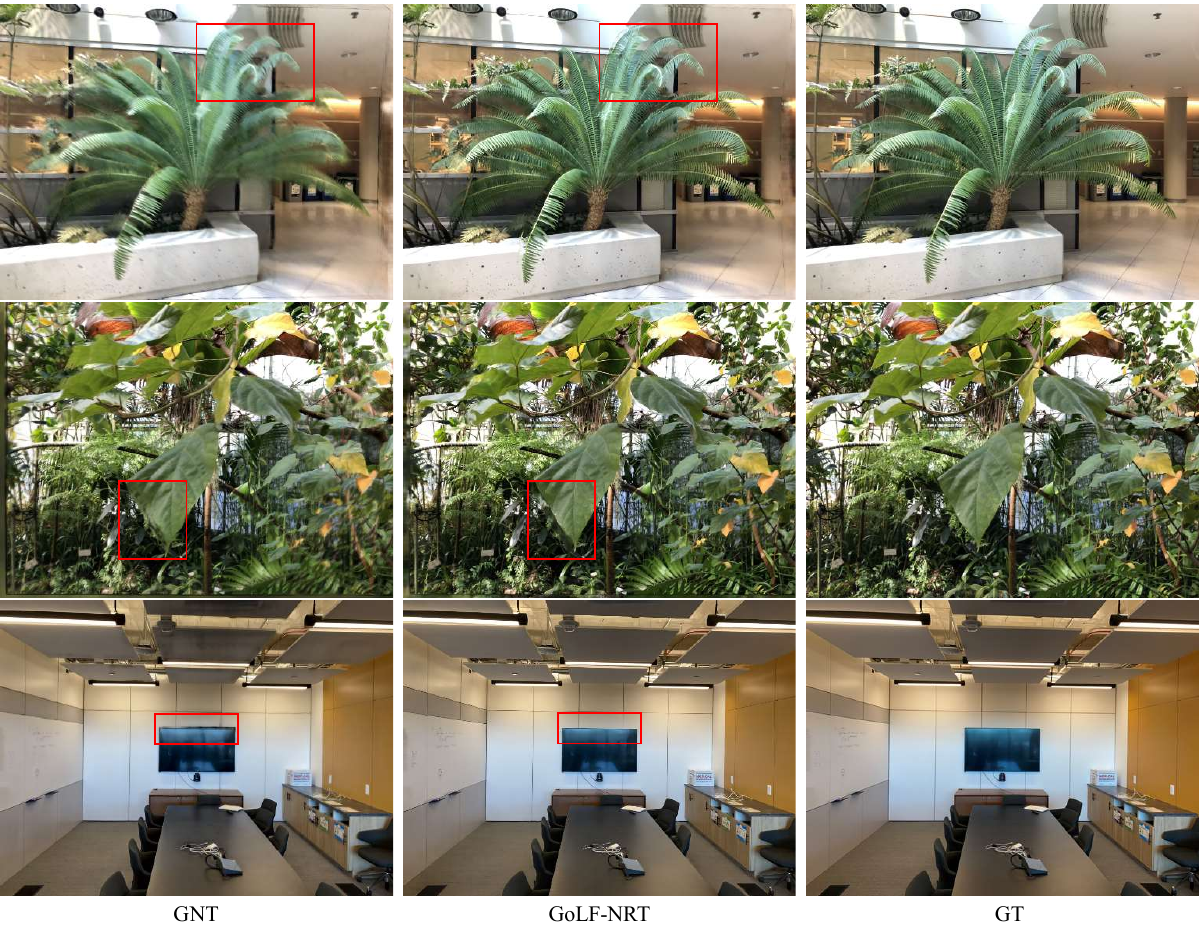} 
\caption{Qualitative comparison between GoLF-NRT and GNT with a single input view. The first, second, and third rows correspond to the Fern, Leaves, and Room scene from LLFF, respectively.}
\label{fig:1per}
\end{figure*}

In this supplementary material, we provide additional details to supplement the main manuscript, encompassing detailed analysis of per-scene optimization experiments and further visual comparisons with existing works. Alongside these, we have also incorporated a demonstration example in GIF format within the supplementary materials, offering a dynamic illustration of our methodology.

\section{Supplementary Notes on Experiments}
\subsection{Visualization Results under a Single Input}
Synthesizing novel views from a single input image represents a highly challenging task, as it constitutes the most extreme scenario among few-shot input conditions. The absence of multi-view information makes obtaining accurate depth information significantly more difficult. As shown in Figure \ref{fig:1view}, inaccuracies in depth estimation can cause misalignments within the synthesized scene, leading to noticeable artifacts. Notably, methods such as GNT \cite{varma2022attention}, which rely heavily on multi-view feature matching, often produce severely distorted rendered images under such conditions. In contrast, our approach introduces global features to enhance the model’s scene comprehension, maintain semantic consistency across objects, and significantly improve the quality and realism of the synthesized views.

\subsection{Results on Reflective/Transparent Surfaces}
Table \ref{tab:per-shiny} summarizes the validation metrics for various methods \cite{varma2022attention, min2024entangled} across five scenarios in the shiny dataset \cite{wizadwongsa2021nex}. Our method consistently achieves the best performance across all scenarios, with particularly significant improvements observed in the CD and Lab scenarios. We attribute this to the increased presence of reflective and transparent surfaces in these scenarios, which pose greater challenges for accurately capturing lighting and geometric information. By incorporating global features, our approach enhances the model’s adaptability to such complexities, enabling a more accurate interpretation of variations in lighting and surface textures. As a result, our method generates more realistic and visually coherent view synthesis outputs, even in challenging conditions.

To further validate our approach, we visualized the rendering results for two scenarios, Materials and Ship, from the Blender dataset \cite{mildenhall2021nerf}, which are representative scenarios featuring reflective surfaces. As shown in Figure \ref{fig:1nerf}, the images rendered by GNT \cite{varma2022attention} exhibit significant loss of geometric detail, leading to blurred boundaries between objects. In contrast, our method produces renderings that more accurately capture the overall appearance and structure of the objects, showcasing superior geometric fidelity and visual clarity.

\section{Per-Scene Optimization Experiments}
\subsection{Implementation Details.}
We conducted fine-tune experiments on each of the eight scenes included in the LLFF dataset: fern, trex, orchids, flowers, etc. To accelerate the training speed, we reduced the number of sampling points from 128 coarse samples and 64 fine samples to 64 coarse samples and 64 fine samples. As neither CaesarNeRF \cite{zhu2023caesarnerf} nor EVE-NeRF \cite{min2024entangled} have open-sourced their experiments related to per-scene optimization experiments, our work is temporarily only compared with GNT \cite{varma2022attention}, our baseline. For each scene, the training process was iterated for 60,000 times.

\subsection{Analysis.}
The detailed experimental results are presented in Table ~\ref{tab:per-scene}. By comparing GoLF with state-of-the-art methods such as GNT \cite{varma2022attention} and several others \cite{mildenhall2019local,mildenhall2021nerf,wizadwongsa2021nex}, GoLF-NRT has achieved favorable performance in most of the evaluated metrics. In terms of PSNR metrics, we have achieved optimal values in more than half of the scenarios, and at least sub-optimal results for the remaining scenarios. Additionally, on the SSIM metric, we have surpassed the majority of existing methods, while we achieved the best experimental results in terms of the LPIPS metric across every scene. Furthermore, compared to GNT \cite{varma2022attention}, our method has consistently yielded superior results across all scenarios and metrics.

\section{Visualizations}
In this section, We present two different variations, framewise results as attached to this document, and the video results in the form of GIF files, which are included in the supplementary material.
\subsection{Framewise Results.} We provide additional examples of per-scene optimization on LLFF, specifically comparing our proposed GoLF-NRT method to GNT \cite{varma2022attention} in terms of both many-shot (i.e., 10 input views) and few-shot (i.e., 1 input views) settings. The results of this comparison are presented in Figure \ref{fig:10per} and Figure \ref{fig:1per}, respectively.

\subsection{Video Results.} In addition to the framewise rendering presented, we have also incorporated rendered videos, in the format of GIF files, as part of the supplementary material accompanying this document. While our primary focus has been on achieving generalizable rendering utilizing few-shot reference views for each frame's reconstruction, for the purpose of video rendering, we showcase examples for two cases, including the rendering results with three reference views for generalizable rendering and per-scene optimization.

In the context of the generalizable setting that utilizes three reference views, we have selectively chosen three scenes from the LLFF dataset characterized by high-frequency pattern variations: "flower", "horns", and "leaves". For these scenes, we conduct a comparative analysis between GoLF-NRT and its baseline method, GNT \cite{varma2022attention}. Our findings indicate that when the input views are limited yet adequate, GNT tends to produce more inconsistent fragments. Moreover, there are a lot of flickering artifacts that are noticeable in the frame. In contrast, our proposed GoLF-NRT, due to the incorporation of more global context information, results in smoother rendered videos, particularly noticeable at the boundaries of leaves and other objects, while also yielding a cleaner overall image.

In the context of the per-scene optimization setting, we present an example involving “orchids”, comparing GoLF-NRT with GNT \cite{varma2022attention}. Similarly, GoLF-NRT produces a more consistent rendering.
\clearpage
\clearpage
{
    \small
    \bibliographystyle{ieeenat_fullname}
    \bibliography{main.bib}

\begin{thebibliography}{70}
\providecommand{\natexlab}[1]{#1}
\providecommand{\url}[1]{\texttt{#1}}
\expandafter\ifx\csname urlstyle\endcsname\relax
  \providecommand{\doi}[1]{doi: #1}\else
  \providecommand{\doi}{doi: \begingroup \urlstyle{rm}\Url}\fi

\bibitem[Bao et~al.(2023)Bao, Li, Huo, Ding, Liang, Li, and Gao]{bao2023and}
Yanqi Bao, Yuxin Li, Jing Huo, Tianyu Ding, Xinyue Liang, Wenbin Li, and Yang Gao.
\newblock Where and how: Mitigating confusion in neural radiance fields from sparse inputs.
\newblock In \emph{Proceedings of the 31st ACM International Conference on Multimedia}, pages 2180--2188, 2023.

\bibitem[Barron et~al.(2021)Barron, Mildenhall, Tancik, Hedman, Martin-Brualla, and Srinivasan]{barron2021mip}
Jonathan~T Barron, Ben Mildenhall, Matthew Tancik, Peter Hedman, Ricardo Martin-Brualla, and Pratul~P Srinivasan.
\newblock Mip-nerf: A multiscale representation for anti-aliasing neural radiance fields.
\newblock In \emph{Proceedings of the IEEE/CVF international conference on computer vision}, pages 5855--5864, 2021.

\bibitem[Barron et~al.(2022)Barron, Mildenhall, Verbin, Srinivasan, and Hedman]{barron2022mip}
Jonathan~T Barron, Ben Mildenhall, Dor Verbin, Pratul~P Srinivasan, and Peter Hedman.
\newblock Mip-nerf 360: Unbounded anti-aliased neural radiance fields.
\newblock In \emph{Proceedings of the IEEE/CVF conference on computer vision and pattern recognition}, pages 5470--5479, 2022.

\bibitem[Barron et~al.(2023)Barron, Mildenhall, Verbin, Srinivasan, and Hedman]{barron2023zip}
Jonathan~T Barron, Ben Mildenhall, Dor Verbin, Pratul~P Srinivasan, and Peter Hedman.
\newblock Zip-nerf: Anti-aliased grid-based neural radiance fields.
\newblock In \emph{Proceedings of the IEEE/CVF International Conference on Computer Vision}, pages 19697--19705, 2023.

\bibitem[Charatan et~al.(2024)Charatan, Li, Tagliasacchi, and Sitzmann]{charatan2024pixelsplat}
David Charatan, Sizhe~Lester Li, Andrea Tagliasacchi, and Vincent Sitzmann.
\newblock pixelsplat: 3d gaussian splats from image pairs for scalable generalizable 3d reconstruction.
\newblock In \emph{Proceedings of the IEEE/CVF Conference on Computer Vision and Pattern Recognition}, pages 19457--19467, 2024.

\bibitem[Chen et~al.(2021)Chen, Xu, Zhao, Zhang, Xiang, Yu, and Su]{chen2021mvsnerf}
Anpei Chen, Zexiang Xu, Fuqiang Zhao, Xiaoshuai Zhang, Fanbo Xiang, Jingyi Yu, and Hao Su.
\newblock Mvsnerf: Fast generalizable radiance field reconstruction from multi-view stereo.
\newblock In \emph{Proceedings of the IEEE/CVF international conference on computer vision}, pages 14124--14133, 2021.

\bibitem[Chen et~al.(2022)Chen, Xu, Geiger, Yu, and Su]{chen2022tensorf}
Anpei Chen, Zexiang Xu, Andreas Geiger, Jingyi Yu, and Hao Su.
\newblock Tensorf: Tensorial radiance fields.
\newblock In \emph{European conference on computer vision}, pages 333--350. Springer, 2022.

\bibitem[Chen et~al.(2023{\natexlab{a}})Chen, Xu, Wu, Zheng, Cham, and Cai]{chen2023explicit}
Yuedong Chen, Haofei Xu, Qianyi Wu, Chuanxia Zheng, Tat-Jen Cham, and Jianfei Cai.
\newblock Explicit correspondence matching for generalizable neural radiance fields.
\newblock \emph{arXiv preprint arXiv:2304.12294}, 2023{\natexlab{a}}.

\bibitem[Chen et~al.(2023{\natexlab{b}})Chen, Funkhouser, Hedman, and Tagliasacchi]{chen2023mobilenerf}
Zhiqin Chen, Thomas Funkhouser, Peter Hedman, and Andrea Tagliasacchi.
\newblock Mobilenerf: Exploiting the polygon rasterization pipeline for efficient neural field rendering on mobile architectures.
\newblock In \emph{Proceedings of the IEEE/CVF Conference on Computer Vision and Pattern Recognition}, pages 16569--16578, 2023{\natexlab{b}}.

\bibitem[Chibane et~al.(2021)Chibane, Bansal, Lazova, and Pons-Moll]{chibane2021stereo}
Julian Chibane, Aayush Bansal, Verica Lazova, and Gerard Pons-Moll.
\newblock Stereo radiance fields (srf): Learning view synthesis for sparse views of novel scenes.
\newblock In \emph{Proceedings of the IEEE/CVF Conference on Computer Vision and Pattern Recognition}, pages 7911--7920, 2021.

\bibitem[Cong et~al.(2023)Cong, Liang, Wang, Fan, Chen, Varma, Wang, and Wang]{cong2023enhancing}
Wenyan Cong, Hanxue Liang, Peihao Wang, Zhiwen Fan, Tianlong Chen, Mukund Varma, Yi Wang, and Zhangyang Wang.
\newblock Enhancing nerf akin to enhancing llms: Generalizable nerf transformer with mixture-of-view-experts.
\newblock In \emph{Proceedings of the IEEE/CVF International Conference on Computer Vision}, pages 3193--3204, 2023.

\bibitem[Deng et~al.(2022)Deng, Liu, Zhu, and Ramanan]{deng2022depth}
Kangle Deng, Andrew Liu, Jun-Yan Zhu, and Deva Ramanan.
\newblock Depth-supervised nerf: Fewer views and faster training for free.
\newblock In \emph{Proceedings of the IEEE/CVF Conference on Computer Vision and Pattern Recognition}, pages 12882--12891, 2022.

\bibitem[Downs et~al.(2022)Downs, Francis, Koenig, Kinman, Hickman, Reymann, McHugh, and Vanhoucke]{downs2022google}
Laura Downs, Anthony Francis, Nate Koenig, Brandon Kinman, Ryan Hickman, Krista Reymann, Thomas~B McHugh, and Vincent Vanhoucke.
\newblock Google scanned objects: A high-quality dataset of 3d scanned household items.
\newblock In \emph{2022 International Conference on Robotics and Automation (ICRA)}, pages 2553--2560. IEEE, 2022.

\bibitem[Flynn et~al.(2019)Flynn, Broxton, Debevec, DuVall, Fyffe, Overbeck, Snavely, and Tucker]{flynn2019deepview}
John Flynn, Michael Broxton, Paul Debevec, Matthew DuVall, Graham Fyffe, Ryan Overbeck, Noah Snavely, and Richard Tucker.
\newblock Deepview: View synthesis with learned gradient descent.
\newblock In \emph{Proceedings of the IEEE/CVF Conference on Computer Vision and Pattern Recognition}, pages 2367--2376, 2019.

\bibitem[Gao et~al.(2023)Gao, Cao, and Shan]{gao2023surfelnerf}
Yiming Gao, Yan-Pei Cao, and Ying Shan.
\newblock Surfelnerf: Neural surfel radiance fields for online photorealistic reconstruction of indoor scenes.
\newblock In \emph{Proceedings of the IEEE/CVF Conference on Computer Vision and Pattern Recognition}, pages 108--118, 2023.

\bibitem[Hedman et~al.(2021)Hedman, Srinivasan, Mildenhall, Barron, and Debevec]{hedman2021baking}
Peter Hedman, Pratul~P Srinivasan, Ben Mildenhall, Jonathan~T Barron, and Paul Debevec.
\newblock Baking neural radiance fields for real-time view synthesis.
\newblock In \emph{Proceedings of the IEEE/CVF international conference on computer vision}, pages 5875--5884, 2021.

\bibitem[Hu et~al.(2023)Hu, Wang, Ma, Yang, Gao, Liu, and Ma]{hu2023tri}
Wenbo Hu, Yuling Wang, Lin Ma, Bangbang Yang, Lin Gao, Xiao Liu, and Yuewen Ma.
\newblock Tri-miprf: Tri-mip representation for efficient anti-aliasing neural radiance fields.
\newblock In \emph{Proceedings of the IEEE/CVF International Conference on Computer Vision}, pages 19774--19783, 2023.

\bibitem[Huang et~al.(2023)Huang, Zhang, Feng, Li, Wang, and Wang]{huang2023local}
Xin Huang, Qi Zhang, Ying Feng, Xiaoyu Li, Xuan Wang, and Qing Wang.
\newblock Local implicit ray function for generalizable radiance field representation.
\newblock In \emph{Proceedings of the IEEE/CVF Conference on Computer Vision and Pattern Recognition}, pages 97--107, 2023.

\bibitem[Jain et~al.(2021)Jain, Tancik, and Abbeel]{jain2021putting}
Ajay Jain, Matthew Tancik, and Pieter Abbeel.
\newblock Putting nerf on a diet: Semantically consistent few-shot view synthesis.
\newblock In \emph{Proceedings of the IEEE/CVF International Conference on Computer Vision}, pages 5885--5894, 2021.

\bibitem[Jin et~al.(2024)Jin, Jiang, Tan, Zhang, Bi, Zhang, Luan, Snavely, and Xu]{jin2024lvsmlargeviewsynthesis}
Haian Jin, Hanwen Jiang, Hao Tan, Kai Zhang, Sai Bi, Tianyuan Zhang, Fujun Luan, Noah Snavely, and Zexiang Xu.
\newblock Lvsm: A large view synthesis model with minimal 3d inductive bias, 2024.

\bibitem[Johari et~al.(2022)Johari, Lepoittevin, and Fleuret]{johari2022geonerf}
Mohammad~Mahdi Johari, Yann Lepoittevin, and Fran{\c{c}}ois Fleuret.
\newblock Geonerf: Generalizing nerf with geometry priors.
\newblock In \emph{Proceedings of the IEEE/CVF Conference on Computer Vision and Pattern Recognition}, pages 18365--18375, 2022.

\bibitem[Kajiya and Von~Herzen(1984)]{kajiya1984ray}
James~T Kajiya and Brian~P Von~Herzen.
\newblock Ray tracing volume densities.
\newblock \emph{ACM SIGGRAPH computer graphics}, 18\penalty0 (3):\penalty0 165--174, 1984.

\bibitem[Kulh{\'a}nek et~al.(2022)Kulh{\'a}nek, Derner, Sattler, and Babu{\v{s}}ka]{kulhanek2022viewformer}
Jon{\'a}{\v{s}} Kulh{\'a}nek, Erik Derner, Torsten Sattler, and Robert Babu{\v{s}}ka.
\newblock Viewformer: Nerf-free neural rendering from few images using transformers.
\newblock In \emph{European Conference on Computer Vision}, pages 198--216. Springer, 2022.

\bibitem[Kwak et~al.(2023)Kwak, Song, and Kim]{kwak2023geconerf}
Min-Seop Kwak, Jiuhn Song, and Seungryong Kim.
\newblock Geconerf: Few-shot neural radiance fields via geometric consistency.
\newblock \emph{arXiv preprint arXiv:2301.10941}, 2023.

\bibitem[Li et~al.(2021)Li, Feng, She, Ding, Wang, and Lee]{li2021mine}
Jiaxin Li, Zijian Feng, Qi She, Henghui Ding, Changhu Wang, and Gim~Hee Lee.
\newblock Mine: Towards continuous depth mpi with nerf for novel view synthesis.
\newblock In \emph{Proceedings of the IEEE/CVF International Conference on Computer Vision}, pages 12578--12588, 2021.

\bibitem[Lin et~al.(2017)Lin, Dollar, Girshick, He, Hariharan, and Belongie]{2017Feature}
Tsung~Yi Lin, Piotr Dollar, Ross Girshick, Kaiming He, Bharath Hariharan, and Serge Belongie.
\newblock Feature pyramid networks for object detection.
\newblock \emph{IEEE Computer Society}, 2017.

\bibitem[Liu et~al.(2021)Liu, Zhang, Zhang, Zhang, Zhu, and Russell]{liu2021editing}
Steven Liu, Xiuming Zhang, Zhoutong Zhang, Richard Zhang, Jun-Yan Zhu, and Bryan Russell.
\newblock Editing conditional radiance fields.
\newblock In \emph{Proceedings of the IEEE/CVF international conference on computer vision}, pages 5773--5783, 2021.

\bibitem[Liu et~al.(2024)Liu, Wang, Hu, Shen, Ye, Zang, Cao, Li, and Liu]{liu2024mvsgaussian}
Tianqi Liu, Guangcong Wang, Shoukang Hu, Liao Shen, Xinyi Ye, Yuhang Zang, Zhiguo Cao, Wei Li, and Ziwei Liu.
\newblock Mvsgaussian: Fast generalizable gaussian splatting reconstruction from multi-view stereo.
\newblock In \emph{European Conference on Computer Vision}, pages 37--53. Springer, 2024.

\bibitem[Liu et~al.(2022)Liu, Peng, Liu, Wang, Wang, Theobalt, Zhou, and Wang]{liu2022neural}
Yuan Liu, Sida Peng, Lingjie Liu, Qianqian Wang, Peng Wang, Christian Theobalt, Xiaowei Zhou, and Wenping Wang.
\newblock Neural rays for occlusion-aware image-based rendering.
\newblock In \emph{Proceedings of the IEEE/CVF Conference on Computer Vision and Pattern Recognition}, pages 7824--7833, 2022.

\bibitem[Mildenhall et~al.(2019)Mildenhall, Srinivasan, Ortiz-Cayon, Kalantari, Ramamoorthi, Ng, and Kar]{mildenhall2019local}
Ben Mildenhall, Pratul~P Srinivasan, Rodrigo Ortiz-Cayon, Nima~Khademi Kalantari, Ravi Ramamoorthi, Ren Ng, and Abhishek Kar.
\newblock Local light field fusion: Practical view synthesis with prescriptive sampling guidelines.
\newblock \emph{ACM Transactions on Graphics (ToG)}, 38\penalty0 (4):\penalty0 1--14, 2019.

\bibitem[Mildenhall et~al.(2021)Mildenhall, Srinivasan, Tancik, Barron, Ramamoorthi, and Ng]{mildenhall2021nerf}
Ben Mildenhall, Pratul~P Srinivasan, Matthew Tancik, Jonathan~T Barron, Ravi Ramamoorthi, and Ren Ng.
\newblock Nerf: Representing scenes as neural radiance fields for view synthesis.
\newblock \emph{Communications of the ACM}, 65\penalty0 (1):\penalty0 99--106, 2021.

\bibitem[Min et~al.(2024)Min, Luo, Yang, Wang, and Yang]{min2024entangled}
Zhiyuan Min, Yawei Luo, Wei Yang, Yuesong Wang, and Yi Yang.
\newblock Entangled view-epipolar information aggregation for generalizable neural radiance fields.
\newblock In \emph{Proceedings of the IEEE/CVF Conference on Computer Vision and Pattern Recognition}, pages 4906--4916, 2024.

\bibitem[Niemeyer et~al.(2022)Niemeyer, Barron, Mildenhall, Sajjadi, Geiger, and Radwan]{niemeyer2022regnerf}
Michael Niemeyer, Jonathan~T Barron, Ben Mildenhall, Mehdi~SM Sajjadi, Andreas Geiger, and Noha Radwan.
\newblock Regnerf: Regularizing neural radiance fields for view synthesis from sparse inputs.
\newblock In \emph{Proceedings of the IEEE/CVF Conference on Computer Vision and Pattern Recognition}, pages 5480--5490, 2022.

\bibitem[Reiser et~al.(2023)Reiser, Szeliski, Verbin, Srinivasan, Mildenhall, Geiger, Barron, and Hedman]{reiser2023merf}
Christian Reiser, Rick Szeliski, Dor Verbin, Pratul Srinivasan, Ben Mildenhall, Andreas Geiger, Jon Barron, and Peter Hedman.
\newblock Merf: Memory-efficient radiance fields for real-time view synthesis in unbounded scenes.
\newblock \emph{ACM Transactions on Graphics (TOG)}, 42\penalty0 (4):\penalty0 1--12, 2023.

\bibitem[Reizenstein et~al.(2021)Reizenstein, Shapovalov, Henzler, Sbordone, Labatut, and Novotny]{reizenstein2021common}
Jeremy Reizenstein, Roman Shapovalov, Philipp Henzler, Luca Sbordone, Patrick Labatut, and David Novotny.
\newblock Common objects in 3d: Large-scale learning and evaluation of real-life 3d category reconstruction.
\newblock In \emph{Proceedings of the IEEE/CVF international conference on computer vision}, pages 10901--10911, 2021.

\bibitem[Roessle et~al.(2022)Roessle, Barron, Mildenhall, Srinivasan, and Nie{\ss}ner]{roessle2022dense}
Barbara Roessle, Jonathan~T Barron, Ben Mildenhall, Pratul~P Srinivasan, and Matthias Nie{\ss}ner.
\newblock Dense depth priors for neural radiance fields from sparse input views.
\newblock In \emph{Proceedings of the IEEE/CVF Conference on Computer Vision and Pattern Recognition}, pages 12892--12901, 2022.

\bibitem[Sajjadi et~al.(2022)Sajjadi, Meyer, Pot, Bergmann, Greff, Radwan, Vora, Lu{\v{c}}i{\'c}, Duckworth, Dosovitskiy, et~al.]{sajjadi2022scene}
Mehdi~SM Sajjadi, Henning Meyer, Etienne Pot, Urs Bergmann, Klaus Greff, Noha Radwan, Suhani Vora, Mario Lu{\v{c}}i{\'c}, Daniel Duckworth, Alexey Dosovitskiy, et~al.
\newblock Scene representation transformer: Geometry-free novel view synthesis through set-latent scene representations.
\newblock In \emph{Proceedings of the IEEE/CVF Conference on Computer Vision and Pattern Recognition}, pages 6229--6238, 2022.

\bibitem[Suhail et~al.(2022{\natexlab{a}})Suhail, Esteves, Sigal, and Makadia]{suhail2022generalizable}
Mohammed Suhail, Carlos Esteves, Leonid Sigal, and Ameesh Makadia.
\newblock Generalizable patch-based neural rendering.
\newblock In \emph{European Conference on Computer Vision}, pages 156--174. Springer, 2022{\natexlab{a}}.

\bibitem[Suhail et~al.(2022{\natexlab{b}})Suhail, Esteves, Sigal, and Makadia]{suhail2022light}
Mohammed Suhail, Carlos Esteves, Leonid Sigal, and Ameesh Makadia.
\newblock Light field neural rendering.
\newblock In \emph{Proceedings of the IEEE/CVF Conference on Computer Vision and Pattern Recognition}, pages 8269--8279, 2022{\natexlab{b}}.

\bibitem[Sun et~al.(2022{\natexlab{a}})Sun, Sun, and Chen]{sun2022direct}
Cheng Sun, Min Sun, and Hwann-Tzong Chen.
\newblock Direct voxel grid optimization: Super-fast convergence for radiance fields reconstruction.
\newblock In \emph{Proceedings of the IEEE/CVF conference on computer vision and pattern recognition}, pages 5459--5469, 2022{\natexlab{a}}.

\bibitem[Sun et~al.(2022{\natexlab{b}})Sun, Wang, Zhang, Li, Zhang, Liu, and Wang]{sun2022fenerf}
Jingxiang Sun, Xuan Wang, Yong Zhang, Xiaoyu Li, Qi Zhang, Yebin Liu, and Jue Wang.
\newblock Fenerf: Face editing in neural radiance fields.
\newblock In \emph{Proceedings of the IEEE/CVF conference on computer vision and pattern recognition}, pages 7672--7682, 2022{\natexlab{b}}.

\bibitem[Sun et~al.(2024)Sun, Wu, and Gao]{sun2024recent}
Jia-Mu Sun, Tong Wu, and Lin Gao.
\newblock Recent advances in implicit representation-based 3d shape generation.
\newblock \emph{Visual Intelligence}, 2\penalty0 (1):\penalty0 9, 2024.

\bibitem[Szymanowicz et~al.(2024)Szymanowicz, Rupprecht, and Vedaldi]{szymanowicz2024splatter}
Stanislaw Szymanowicz, Chrisitian Rupprecht, and Andrea Vedaldi.
\newblock Splatter image: Ultra-fast single-view 3d reconstruction.
\newblock In \emph{Proceedings of the IEEE/CVF Conference on Computer Vision and Pattern Recognition}, pages 10208--10217, 2024.

\bibitem[Takeda et~al.(2005)Takeda, Farsiu, and Milanfar]{2005Image}
Hiroyuki Takeda, Sina Farsiu, and Peyman Milanfar.
\newblock Image denoising by adaptive kernel regression.
\newblock \emph{Circuits Systems \& Computers .conference Record.asilomar Conference on}, 2005, 2005.

\bibitem[Takeda et~al.(2006)Takeda, Farsiu, and Milanfar]{2006Robust}
Hiroyuki Takeda, Sina Farsiu, and Peyman Milanfar.
\newblock Robust kernel regression for restoration and reconstruction of images from sparse noisy data.
\newblock \emph{IEEE}, 2006.

\bibitem[Takeda et~al.(2007)Takeda, Farsiu, and Milanfar]{2007Kernel}
H. Takeda, S. Farsiu, and P. Milanfar.
\newblock Kernel regression for image processing and reconstruction.
\newblock \emph{IEEE Transactions on Image Processing}, 2007.

\bibitem[Trevithick and Yang(2021)]{trevithick2021grf}
Alex Trevithick and Bo Yang.
\newblock Grf: Learning a general radiance field for 3d representation and rendering.
\newblock In \emph{Proceedings of the IEEE/CVF International Conference on Computer Vision}, pages 15182--15192, 2021.

\bibitem[Tu et~al.(2022)Tu, Talebi, Zhang, Yang, and Li]{2022MaxViT}
Zhengzhong Tu, Hossein Talebi, Han Zhang, Feng Yang, and Yinxiao Li.
\newblock Maxvit: Multi-axis vision transformer.
\newblock \emph{arXiv e-prints}, 2022.

\bibitem[Varma et~al.(2022)Varma, Wang, Chen, Chen, Venugopalan, and Wang]{varma2022attention}
Mukund Varma, Peihao Wang, Xuxi Chen, Tianlong Chen, Subhashini Venugopalan, and Zhangyang Wang.
\newblock Is attention all that nerf needs?
\newblock In \emph{The Eleventh International Conference on Learning Representations}, 2022.

\bibitem[Vaswani et~al.(2017)Vaswani, Shazeer, Parmar, Uszkoreit, Jones, Gomez, Kaiser, and Polosukhin]{vaswani2017attention}
Ashish Vaswani, Noam Shazeer, Niki Parmar, Jakob Uszkoreit, Llion Jones, Aidan~N Gomez, {\L}ukasz Kaiser, and Illia Polosukhin.
\newblock Attention is all you need.
\newblock \emph{Advances in neural information processing systems}, 30, 2017.

\bibitem[Venkat et~al.(2023)Venkat, Agarwal, Singh, and Tulsiani]{venkat2023geometry}
Naveen Venkat, Mayank Agarwal, Maneesh Singh, and Shubham Tulsiani.
\newblock Geometry-biased transformers for novel view synthesis.
\newblock \emph{arXiv preprint arXiv:2301.04650}, 2023.

\bibitem[Wang et~al.(2022{\natexlab{a}})Wang, Chai, He, Chen, and Liao]{wang2022clip}
Can Wang, Menglei Chai, Mingming He, Dongdong Chen, and Jing Liao.
\newblock Clip-nerf: Text-and-image driven manipulation of neural radiance fields.
\newblock In \emph{Proceedings of the IEEE/CVF Conference on Computer Vision and Pattern Recognition}, pages 3835--3844, 2022{\natexlab{a}}.

\bibitem[Wang et~al.(2022{\natexlab{b}})Wang, Cui, Salcudean, and Wang]{wang2022generalizable}
Dan Wang, Xinrui Cui, Septimiu Salcudean, and Z~Jane Wang.
\newblock Generalizable neural radiance fields for novel view synthesis with transformer.
\newblock \emph{arXiv preprint arXiv:2206.05375}, 2022{\natexlab{b}}.

\bibitem[Wang et~al.(2023)Wang, Chen, Loy, and Liu]{wang2023sparsenerf}
Guangcong Wang, Zhaoxi Chen, Chen~Change Loy, and Ziwei Liu.
\newblock Sparsenerf: Distilling depth ranking for few-shot novel view synthesis.
\newblock In \emph{Proceedings of the IEEE/CVF International Conference on Computer Vision}, pages 9065--9076, 2023.

\bibitem[Wang et~al.(2021)Wang, Wang, Genova, Srinivasan, Zhou, Barron, Martin-Brualla, Snavely, and Funkhouser]{wang2021ibrnet}
Qianqian Wang, Zhicheng Wang, Kyle Genova, Pratul~P Srinivasan, Howard Zhou, Jonathan~T Barron, Ricardo Martin-Brualla, Noah Snavely, and Thomas Funkhouser.
\newblock Ibrnet: Learning multi-view image-based rendering.
\newblock In \emph{Proceedings of the IEEE/CVF conference on computer vision and pattern recognition}, pages 4690--4699, 2021.

\bibitem[Wang et~al.(2004)Wang, Bovik, Sheikh, and Simoncelli]{wang2004image}
Zhou Wang, Alan~C Bovik, Hamid~R Sheikh, and Eero~P Simoncelli.
\newblock Image quality assessment: from error visibility to structural similarity.
\newblock \emph{IEEE transactions on image processing}, 13\penalty0 (4):\penalty0 600--612, 2004.

\bibitem[Wizadwongsa et~al.(2021)Wizadwongsa, Phongthawee, Yenphraphai, and Suwajanakorn]{wizadwongsa2021nex}
Suttisak Wizadwongsa, Pakkapon Phongthawee, Jiraphon Yenphraphai, and Supasorn Suwajanakorn.
\newblock Nex: Real-time view synthesis with neural basis expansion.
\newblock In \emph{Proceedings of the IEEE/CVF Conference on Computer Vision and Pattern Recognition}, pages 8534--8543, 2021.

\bibitem[Xiang et~al.(2021)Xiang, Xu, Hasan, Hold-Geoffroy, Sunkavalli, and Su]{xiang2021neutex}
Fanbo Xiang, Zexiang Xu, Milos Hasan, Yannick Hold-Geoffroy, Kalyan Sunkavalli, and Hao Su.
\newblock Neutex: Neural texture mapping for volumetric neural rendering.
\newblock In \emph{Proceedings of the IEEE/CVF Conference on Computer Vision and Pattern Recognition}, pages 7119--7128, 2021.

\bibitem[Xu et~al.(2022)Xu, Jiang, Wang, Fan, Shi, and Wang]{xu2022sinnerf}
Dejia Xu, Yifan Jiang, Peihao Wang, Zhiwen Fan, Humphrey Shi, and Zhangyang Wang.
\newblock Sinnerf: Training neural radiance fields on complex scenes from a single image.
\newblock In \emph{European Conference on Computer Vision}, pages 736--753. Springer, 2022.

\bibitem[Xu et~al.(2024)Xu, Chen, Chen, Sakaridis, Zhang, Pollefeys, Geiger, and Yu]{xu2024murf}
Haofei Xu, Anpei Chen, Yuedong Chen, Christos Sakaridis, Yulun Zhang, Marc Pollefeys, Andreas Geiger, and Fisher Yu.
\newblock Murf: Multi-baseline radiance fields.
\newblock In \emph{Proceedings of the IEEE/CVF Conference on Computer Vision and Pattern Recognition}, pages 20041--20050, 2024.

\bibitem[Xu and Harada(2022)]{xu2022deforming}
Tianhan Xu and Tatsuya Harada.
\newblock Deforming radiance fields with cages.
\newblock In \emph{European Conference on Computer Vision}, pages 159--175. Springer, 2022.

\bibitem[Yang et~al.(2023{\natexlab{a}})Yang, Hong, Li, Hu, Li, Lee, and Wang]{yang2023contranerf}
Hao Yang, Lanqing Hong, Aoxue Li, Tianyang Hu, Zhenguo Li, Gim~Hee Lee, and Liwei Wang.
\newblock Contranerf: Generalizable neural radiance fields for synthetic-to-real novel view synthesis via contrastive learning.
\newblock In \emph{Proceedings of the IEEE/CVF Conference on Computer Vision and Pattern Recognition}, pages 16508--16517, 2023{\natexlab{a}}.

\bibitem[Yang et~al.(2023{\natexlab{b}})Yang, Pavone, and Wang]{yang2023freenerf}
Jiawei Yang, Marco Pavone, and Yue Wang.
\newblock Freenerf: Improving few-shot neural rendering with free frequency regularization.
\newblock In \emph{Proceedings of the IEEE/CVF conference on computer vision and pattern recognition}, pages 8254--8263, 2023{\natexlab{b}}.

\bibitem[Yu et~al.(2021{\natexlab{a}})Yu, Li, Tancik, Li, Ng, and Kanazawa]{yu2021plenoctrees}
Alex Yu, Ruilong Li, Matthew Tancik, Hao Li, Ren Ng, and Angjoo Kanazawa.
\newblock Plenoctrees for real-time rendering of neural radiance fields.
\newblock In \emph{Proceedings of the IEEE/CVF International Conference on Computer Vision}, pages 5752--5761, 2021{\natexlab{a}}.

\bibitem[Yu et~al.(2021{\natexlab{b}})Yu, Ye, Tancik, and Kanazawa]{yu2021pixelnerf}
Alex Yu, Vickie Ye, Matthew Tancik, and Angjoo Kanazawa.
\newblock pixelnerf: Neural radiance fields from one or few images.
\newblock In \emph{Proceedings of the IEEE/CVF conference on computer vision and pattern recognition}, pages 4578--4587, 2021{\natexlab{b}}.

\bibitem[Zhang et~al.(2021)Zhang, Yang, Tulsiani, and Ramanan]{zhang2021ners}
Jason Zhang, Gengshan Yang, Shubham Tulsiani, and Deva Ramanan.
\newblock Ners: Neural reflectance surfaces for sparse-view 3d reconstruction in the wild.
\newblock \emph{Advances in Neural Information Processing Systems}, 34:\penalty0 29835--29847, 2021.

\bibitem[Zhang et~al.(2018)Zhang, Isola, Efros, Shechtman, and Wang]{zhang2018unreasonable}
Richard Zhang, Phillip Isola, Alexei~A Efros, Eli Shechtman, and Oliver Wang.
\newblock The unreasonable effectiveness of deep features as a perceptual metric.
\newblock In \emph{Proceedings of the IEEE conference on computer vision and pattern recognition}, pages 586--595, 2018.

\bibitem[Zheng et~al.(2024)Zheng, Zhou, Shao, Liu, Zhang, Nie, and Liu]{zheng2024gps}
Shunyuan Zheng, Boyao Zhou, Ruizhi Shao, Boning Liu, Shengping Zhang, Liqiang Nie, and Yebin Liu.
\newblock Gps-gaussian: Generalizable pixel-wise 3d gaussian splatting for real-time human novel view synthesis.
\newblock In \emph{Proceedings of the IEEE/CVF Conference on Computer Vision and Pattern Recognition}, pages 19680--19690, 2024.

\bibitem[Zhou et~al.(2018)Zhou, Tucker, Flynn, Fyffe, and Snavely]{zhou2018stereo}
Tinghui Zhou, Richard Tucker, John Flynn, Graham Fyffe, and Noah Snavely.
\newblock Stereo magnification: Learning view synthesis using multiplane images.
\newblock \emph{arXiv preprint arXiv:1805.09817}, 2018.

\bibitem[Zhu et~al.(2024)Zhu, Ding, Chen, Zharkov, Nevatia, and Liang]{zhu2023caesarnerf}
Haidong Zhu, Tianyu Ding, Tianyi Chen, Ilya Zharkov, Ram Nevatia, and Luming Liang.
\newblock Caesarnerf: Calibrated semantic representation for few-shot generalizable neural rendering.
\newblock In \emph{European Conference on Computer Vision}, 2024.

\end{thebibliography}
}

\include{_rebuttal}
\end{document}